%% file: main.tex
\documentclass[sigconf]{acmart}
\AtBeginDocument{%
  \providecommand\BibTeX{{%
    \normalfont B\kern-0.5em{\scshape i\kern-0.25em b}\kern-0.8em\TeX}}}

\setcopyright{acmcopyright}
\copyrightyear{2021}
\acmYear{2021}
\acmDOI{X}
\acmPrice{15.00}
\acmISBN{978-X-XXXX-XXXX-X/YY/MM}

\usepackage{graphicx}
\usepackage{caption,subcaption}
\usepackage{amsfonts}
\usepackage{amsmath}
\usepackage{algorithm}
\usepackage{algorithmicx}
\usepackage[noend]{algpseudocode}
\usepackage{url}
\usepackage{multirow}
\usepackage{bm}

\newtheorem{theorem}{Theorem}

\graphicspath{ {figures/} {diagrams/} {individuals/}}

\begin{document}

\title{Equilibrium Inverse Reinforcement Learning \\ for Ride-hailing Vehicle Network}

\author{Takuma Oda}
\email{takuma.oda@mo-t.com}
\affiliation{%
  \institution{Mobility Technologies Co., Ltd.}
%   \streetaddress{3-12 Kioicho}
%   \city{Chiyoda City}
%   \state{Tokyo, Japan}
%   \postcode{102-0094}
}
\begin{abstract}
Ubiquitous mobile computing have enabled ride-hailing services to collect vast amounts of behavioral data of riders and drivers and optimize supply and demand matching in real time. 
While these mobility service providers have some degree of control over the market by assigning vehicles to requests, they need to deal with the uncertainty arising from self-interested driver behavior since workers are usually free to drive when they are not assigned tasks.
% While these mobility service providers have some degree of control over the market by assigning vehicles to requests, workers are usually free to drive when they are not assigned tasks, and thus need to deal with the uncertainty arising from self-interested driver behavior. 
If a driver's behavior can be accurately replicated on the digital twin, more detailed and realistic counterfactual simulations will enable decision making to improve mobility services as well as to validate urban planning.

In this work, we formulate the problem of passenger-vehicle matching in a sparsely connected graph and proposed an algorithm to derive an equilibrium policy in a multi-agent environment. Our framework combines value iteration methods to estimate the optimal policy given expected state visitation and policy propagation to compute multi-agent state visitation frequencies. Furthermore, we developed a method to learn the driver's reward function transferable to an environment with significantly different dynamics from training data. We evaluated the robustness to changes in spatio-temporal supply-demand distributions and deterioration in data quality using a real-world taxi trajectory dataset; our approach significantly outperforms several baselines in terms of imitation accuracy. The computational time required to obtain an equilibrium policy shared by all vehicles does not depend on the number of agents, and even on the scale of real-world services, it takes only a few seconds on a single CPU.

% Ride-hailing services have been growing rapidly around the world by allowing the collection of vast amounts of behavioral log data of riders and drivers for efficient supply-demand matching. These mobility service providers can intervene in the market by dispatching, while they need to address uncertainty arisen from behaviors of self-interested drivers since they are usually free to drive when a task is not assigned. If a driver's behavior can be accurately replicated on the digital twin, more detailed and realistic counterfactual simulations will enable us to make decisions to improve their services as well as to validate urban planning.

\end{abstract}

\begin{CCSXML}
<ccs2012>
   <concept>
       <concept_id>10002951.10003227.10003236</concept_id>
       <concept_desc>Information systems~Spatial-temporal systems</concept_desc>
       <concept_significance>300</concept_significance>
       </concept>
   <concept>
       <concept_id>10010147.10010341</concept_id>
       <concept_desc>Computing methodologies~Modeling and simulation</concept_desc>
       <concept_significance>300</concept_significance>
       </concept>
 </ccs2012>
\end{CCSXML}

\ccsdesc[300]{Information systems~Spatial-temporal systems}
\ccsdesc[300]{Computing methodologies~Modeling and simulation}

\keywords{Counterfactual simulation, inverse reinforcement learning, Markov games, equilibrium}

\maketitle

\section{Introduction}\label{sec:introduction}
\input{intro}

\section{Related Work}\label{sec:related}
\input{related}

\section{Problem Definition}\label{sec:problem}
\input{problem}

\section{Framework}\label{sec:framework}
\input{framework}

\section{Experiments}\label{sec:experiments}
\input{experiments}

\section{Results and Discussion}\label{sec:results}
\input{results}

\section{Conclusions}\label{sec:conclusions}
\input{conclusions}

\bibliographystyle{ACM-Reference-Format}
\bibliography{reference}

\clearpage

\appendix
\section{Appendix}\label{sec:appendix}
\input{appendix}

\end{document}

%% file: intro.tex
With the ubiquitous use of Internet connected mobile devices which collect data and provide computation in real time, mobility services have been seamlessly linking all types of public transportation, from buses, trains, and taxis and shared bicycles, in order to make it more efficient and convenient. One of the key enablers of this revolution lies in the realistic and accurate prediction and simulation of human decision making.
For ride-hailing platforms such as Uber and Lyft, where service quality depends heavily on drivers behavior, simulation is also an essential tool for testing route optimization and efficiency of order dispatch. In particular, since it is necessary to address uncertainty arisen from behaviors of uncoordinated self-interested drivers, the use of accurate driver behavior models enables more detailed counterfactual simulations, which will be beneficial for making decisions to improve services.

% As mobility becomes increasingly digital, it is becoming possible to create a more convenient and sustainable lifestyle than that of the private car, combining public transport, taxis, shared bicycles and other complex mobility services.
% Realistic and accurate simulation technologies are becoming increasingly important as it is possible to communicate with mobility users who have been offline until now.
% For ride-hailing services such as Uber and Lyft, simulation is also essential for testing route optimization and efficiency of dispatch matching. In particular, if we can accurately simulate driver behavior based on context (e.g., season), we can simulate counterfactual situations such as changes in rider behavior patterns and the number of vehicles in the market, which is useful for making business policy decisions.

Imitation of multi-vehicle movement on sparsely connected road graph poses several significant computational challenges.
One such challenge is the effective modeling of the graph structures in geographical space. A road network graph is often abstracted and represented as a coarser mesh, since finer-grained data structures tend to be more computationally intensive. However, more sophisticated simulation requires reproduction of road-level driving behavior, which will allow us to analyze vehicle routing to passengers and traffic congestion. 
In addition, the number of vehicles to be simulated for a ride-hailing service is generally relatively large, ranging from hundreds to thousands. In order to utilize simulation of this scale for actual decision making, the simulation system must be reasonable in terms of execution time and infrastructure cost.
% One such challenge is the use of the graph structures in geographical space, which is are often abstracted and represented as a coarser mesh. However, more sophisticated simulation using road network will allow us to analyze vehicle routing to passengers and traffic congestion. 
Another requirement for practical use is robustness to changes in environmental dynamics and data noise, which is essential when you want to simulate situations that are not in the distribution of training data. To our knowledge, there are no existing literature dealing with these challenges in ride-hailing vehicle network. 
% In addition, the number of vehicles to be simulated for a ride-hailing service is usually several hundred to several thousand. In order to utilize simulation of this scale for actual decision making, the simulation system must be reasonable in terms of execution time and infrastructure cost.

% As a result, the ability to efficiently generate the actions of agents in a simulation is a necessary metric.

In this work, we propose SEIRL (spatial equilibrium inverse reinforcement learning), the first approach to multi-agent behavioral modeling and reward learning in equilibrium on a road network for a ride-hailing service. We focus on a fleet network where each driver is uncooperatively searching for passengers on streets. We propose MDP formulation of passenger-seeking drivers in a road network and a method to compute the optimal policy shared between agents which enables efficient and scalable simulation. We demonstrate that SEIRL is able to recover reward functions that are robust to significant change in the environment seen during training.

% , allowing us to learn policies even under significant variation in the environment seen during training.
% Thus, we take a \emph{distributed approach} in which each vehicle solves its own DQN problem, without coordination.
% MOVI uses a reinforcement learning technique called deep Q-network (DQN)~\cite{DL,DQN} that focuses on finding the optimal actions rather than accurately modeling the system.

% \subsection{Related Work}
% \input{related}

% \subsection{Our Contributions}
This paper makes the following three major contributions:
\begin{itemize}
    \item \textbf{Vehicle behavior modeling on the road network:} 
    To the best of our knowledge, this is the first study to design multi-agent behavioral modeling in equilibrium on a road network for a ride-hailing service. We have formulated a mathematical model of the ride-hailing matching problem in Markov Decision Processes and proposed an algorithm, Equilibrium Value Iteration, to derive an optimal policy for multiple vehicles in equilibrium, which is efficient in a sense that the policy is shared between agents.
    \item \textbf{Spatial Equilibrium Inverse Reinforcement Learning:} Within the MaxEnt IRL framework, we integrate SEIRL, an algorithm for learning driver rewards and policies in equilibrium from real multi-vehicle trajectory data. We derive the gradient of MaxEnt IRL when reward changes are accompanied by a change in dynamics and demonstrate that the algorithm is stable under certain conditions.
    \item \textbf{Data-driven Evaluation and Counterfactual Simulation:} We compared and validated the robustness of SEIRL to changes in dynamics and data noise using real taxi trajectory data in Yokohama City and showed that it obtain significant performance gains over several baselines with unknown dynamics. In addition, we demonstrated the value of counterfactual simulations for supply vehicle optimization given the objective function of the platform.
\end{itemize}

%% file: related.tex
Imitation learning, such as behavior cloning \cite{pomerleau1991efficient} and generative adversarial imitation learning \cite{ho2016generative}, is a powerful approach to learning policies directly from expert demonstrations to produce a expert-like behavior. They, however, are typically incapable of being applied to general transfer setting. Inferring reward functions from expert demonstrations, which we refer to as inverse reinforcement learning (IRL), is more beneficial for application of counterfactual simulations since the reward function is often considered as the most succinct, robust and transferable representation of a task \cite{abbeel2004apprenticeship} \cite{fu2017learning}. However, IRL is fundamentally ill-defined: in general, many reward functions can make behavior seem rational. To solve the ambiguity, Maximum entropy inverse reinforcement learning (MaxEnt IRL) \cite{ziebart2008maximum} provides a general probabilistic framework by finding the reward function that matches trajectory distribution of the experts. 
Ziebart et.al has demonstrated an approach to simulate behavior on a road network by learning the routing costs of occupied taxi driving using MaxEntIRL \cite{ziebart2008navigate} in single agent setting.

In multi-agent settings, an agent cannot myopically maximize its reward; it must speculate on how the other agents may act to influence the game’s outcome. Multi-agent systems in which the set of agents behave non-cooperatively according to their own interests are modeled by Markov games. Nash equilibrium \cite{hu1998multiagent} is stable outcomes in Markov games where each agent acts optimally with respect to the actions of the other agents. However, Nash equilibrium, in the sense that it assumes that real behavior is typically consistently optimal or completely rational, is incompatible with the MaxEnt IRL framework.
A recent study extends MaxEnt IRL to Markov games with logistic quantal response equilibrium, proposing model-free IRL solution in multi-agent environment \cite{yu2019multi}.
% proposed a new solution concept combining with logistic quantal response equilibrium and Gibbs sampling, for model-free IRL in multi-agent environment.

This work relates to multi-agent inverse reinforcement learning (MAIRL), which includes two-player zero-sum games\cite{lin2017multiagent}, two-player general-sum stochastic game\cite{lin2019multi}, and cooperative games\cite{natarajan2010multi}, in which multiple agent act independently to achieve a common goal.
Waugh et.al introduced a technique for predicting and generalizing behavior in competitive and cooperative multi-agent domains, employing the game-theoretic notion of regret and the principle of maximum entropy \cite{waugh2013computational}. Although the proposed method, Dual MaxEnt ICE, is performed in computation time that scales polynomially in the number of decision makers, it is still limited to small size games, and not applicable to games of large number of players such as ride-hailing services.

We concern ourselves with Markov potential games, a model settings in which agents compete for a common resource such as networking games in telecommunications \cite{altman2006survey}. Transportation network is also a classical game theory problem: each agent seeks to minimize her own individual cost by choosing the best route to reach her destination without taking into account the overall utility. It has been widely studied that selfish routing results in a Nash Equilibrium and often society has to pay a price of anarchy due to lack of coordination \cite{youn2008price}. In contrast, our goal is to develop robust imitation algorithms in ride-hailing services where drivers do not have clear destination and move more stochasticly in search of passengers. Some studies address driver relocation problem for supply and demand matching in continuous geographical state space. Mguni et.al devise a learning method of Nash equilibrium policies that works when the number of agents is extremely large using mean field game framework \cite{mguni2018decentralised} \cite{mguni2019coordinating}.

Related to our study, there are many works on fleet management for ride-hailing platform and connected taxi cabs. For instance, Xu et.al designed an order dispatch algorithm to optimize resource utilization and user experience in a global and farsighted view \cite{xu2018large}. Li et.al proposed MARL solutions for order dispatching which supports fully distributed execution \cite{li2019efficient}. Others have proposed vehicle re-balancing methods for autonomous vehicles \cite{zhang2016control}, considering both global service fairness and future costs. Most of these works ignore the complexity of the fine-grained road network and represent the geographical space as a coarser mesh. This reduces the number of states and allows for simple data structures to be used, which dramatically increases processing speed. By treating supply-demand distribution as an image, recent studies have taken advantage of model-free reinforcement learning to fit optimal relocation policy \cite{oda2018movi} \cite{he2019spatio}. While these methods have mostly been validated in coarse-grained spaces as small as one kilometer square, they are not applicable to behavioral models in a sparsely connected road graph. Some studies have done to optimize routing on a road network: passenger-seeking route recommendation for taxi drivers \cite{yuan2012t} \cite{qu2014cost} and multi-agent routing for mapping using a graph neural network based model \cite{sykora2020multi}. Although these approaches can generate behavioral sequences that reflect the context of each agent, they require heavy inference processing for each agent. This is not practical for large-scale simulations of mobility services that simulate thousands of agents traveling over a road network with tens of thousands of nodes.
% However, to our knowledge, there are no examples that mimic the behavior of empty taxis, which are more complex than a ride with unclear destinations.  
% In recent years, GCNs have been successfully used to represent optimal behavioral sequences that take into account multi-agent interactions \cite{sykora2020multi}; while the generation of behavioral sequences using GCNs can reflect per-agent contexts, it requires heavy inference processing on a per-agent basis and requires tens of thousands to tens of This is not realistic for large-scale simulations of mobility services that simulate thousands of agents moving on a 10,000-node road network. they all require to execute route search algorithm for each driver.

% superhuman poker AI \cite{brown2019superhuman}

%% file: problem.tex
Our goal to imitate passenger-seeking behaviors of multiple taxis in a road network with unknown dynamics. We aim to simulate agents even under significant variation in the environment seen during training. To simplify the problem, we focus on modeling drivers searching for passengers on a street, not drivers waiting for ride-hailing app requests.
% given environmental parameters in a certain context (e.g., OD distribution of passengers, number of vehicles, etc.), if it were the same context in reality. To simplify the problem, we focused on simulating vehicles intended to ride a hailing passenger on streets, not a ride-hailing app user. We are interested in a counterfactual simulation, requiring to imitate behaviors of vehicles in a context which real behavior logs for training does not include.

Formally, the taxi-passenger matching problem is defined by $N$ player Markov games which characterized the following components. 

\textbf{State} of the Markov game at time step $t$ is denoted by $S_t$, which includes the geographical location $s_t^i$ and empty status $o_t^i$ of individual agent $i$. The agent position $s \in \mathcal{S}$ represents a road segment, which corresponds to a node in the strongly connected directed graph $G(\mathcal{S}, \mathcal{A})$ representing the road connectivity. $o$ is a binary variable, where $o = 0$ denotes that the agent is vacant and $o = 1$ denotes that the agent is occupied. 

% Note that we do not incorporate real-world date time into a state since the environmental parameters are assumed to be static.

\textbf{Action}, $a_t^i$ stands for the edge to the next road that agent $i$ follows. The edge $a \in \mathcal{A}$ connects the road network's predecessor node $s$ to the successor node $s'$. We use bold variables without subscript $i$ to denote the concatenation of all variables for all agents (e.g., $\bm{a}=(a^1,...,a^N)$ denotes actions of all agents). Executing $a$ causes a matching with passengers with probability $\rho_a$ and state transition subject to the destination distribution of passengers $d_{s,s'}^t$. This ride trip takes a time step of $h_{s,s'}^t$ which depends on the road network distance between nodes $s$ and $s'$, leading to a transition to $s_{t + h_{s, s'}^t}^i=s'$. On the other hand, if no matching occurs, the agent transitions to $s_{t+\tau_{a}^t}^i=s'$, with a move from the beginning of the road $s$ to the beginning of the road $s'$, which requires a time step of $\tau_{a}^t$. Here, we use a mathematical model endowed with temporally extended actions which are also known as options and corresponding decision problem for single-agent is known as a semi-Markov decision process \cite{sutton1999between} \cite{tang2019deep}. In a multi-agent setting, since each agent interacts with an environment asynchronously, a state of the environment needs to be determined at every time step. Thus, we assume that an agent remains in the occupied state $o_t^i=1$ while carrying a passenger and no actions to take until it arrived the passenger's destination. Similarly, a vacant agent traveling on a road is not allowed to take any action until it arrived at next node.
%  is the fee obtained for finding a customer at time $t$ subtracting the cost of travel due to the action $a_t^i$

\textbf{Reward} of agent $i$ is denoted as $r_t^i$. Each action $a_t^i$ takes a moving cost depending on various factors such as driver preferences. If an agent finds a passenger during executing the action, it receives a reward $w_{s,s''}$ based on the ride trip time. We assume that agents are not cooperative and the reward function is not shared, but identical for all agents.

\textbf{Policy}, $\bm{\pi}(\bm{a}|S)=\prod_{i=1}^N \pi^i(a^i|S)$ specifies the joint probability of taking action $\bm{a}$ in state $S$. We require agents' actions in each state to be independent. All agents act to maximize the same objective by executing actions through a stochastic policy $\pi^i$. The objective of each agent $i$ is maximizing its own expected return $\mathbf{E}[\sum_t\gamma^t r_t^i| S_0, \bm{\pi}]$ where $\gamma$ is the discount factor and $S_0$ is the initial state of all agents. We further define expected return $R^{\pi}(S_t,a_t^i) = \mathbf{E}[\sum_{t'=t}\gamma^{t'-t}r_t^i|S_t,a_t^i,\bm{\pi}]$. To take bounded rationality into consideration, we introduce logistic quantal response equilibrium as a stochastic generalization to Nash equilibrium, where for each state and action, the constraint is given by \cite{yu2019multi} \cite{mckelvey1995quantal}:
% $\bm{\mu_0}(S)=\prod_{i=1}^N \mu_0(s_0^i)$

\begin{equation}\label{eq:lqre}
\pi^i(a^i|S) = \frac{\exp(R^{\pi}(S,a^i))}{\sum_{a^i}\exp(R^{\pi}(S,a^i)}
\end{equation}

Since the actual reward function is unknown, we consider the problem of inverse reinforcement learning, which estimates the reward function from the driver's behavioral trajectory. Function approximation is used in order to regress the feature representation of a state-action onto a real valued reward using a mapping $r = g_{\theta}(s, a)$. Starting from the initial state distribution $p(S_0)$, agents generate a trajectory executing joint policy $\bm{\pi}$ in the environment, denoted as $\{ \bm{\zeta}_j = ((S_{j,0},\bm{a}_{j,0}),...,(S_{j,T},\bm{a}_{j,T})) \}_{j \in \mathcal{J}}$ where $\mathcal{J}$ stands for the index set of the historical trajectories.
Inverse reinforcement learning can be interpreted as estimating the parameters of the reward function $\theta$ by solving the following maximum likelihood estimation problem for the expert's action trajectory $\bm{\zeta}$ in the given data set $\mathcal{D}$:

\begin{equation}
\begin{gathered}\label{eq:ma-irl}
\max_{\theta} L(\theta) = \max_{\theta} \mathbf{E}_{\zeta \sim \mathcal{D}} [\log p_{\theta}(\bm{\zeta})] \\
p_{\theta}(\bm{\zeta}) = p(S_0) \prod_{t=0}^{T}p(S_{t+1}|S_t,\bm{a_t})\prod_{i=1}^N \pi_{\theta}^i(a_t^i|S_t)
\end{gathered}
\end{equation}

where $\pi_{\theta}^i(a_t^i|S_t))$ are the unique stationary distributions for the equilibrium in Equation (\ref{eq:lqre}) induced by $g_{\theta}$.

% Consider estimating a behavior policy when the driver's reward function is known. In the reinforcement learning, the problem is to find the policy that maximizes the driver's cumulative reward as follows

% \begin{equation}\label{eq:rl}
% \pi^* = \text{arg}\max_{\pi}  \text{E}_{s \sim \mu, \tau \sim \pi}[\sum _t \gamma^t r_t|s_0=s] = \text{arg}\max_{\pi}  \text{E}_{s \sim \mu} [V_{\pi}(s)] .
% \end{equation}

% \begin{equation}
% \begin{gathered}\label{eq:rl}
% \pi^* = \text{arg}\max_{\pi}  \text{E}_{s \sim \mu, \tau \sim \pi}[\sum _t \gamma^t r(s_t, a_t)|s_0=s] = \text{arg}\max_{\pi}  \text{E}_{s \sim \mu} [V_{\pi}(s)] \\
% V_{\pi}(s) = \text{E}_{\tau \sim \pi}[\sum _t \gamma^t r(s_t, a_t)|s_0=s] \\
% Q_{\pi}(s, a) = \text{E}_{\tau \sim \pi}[\sum _t \gamma^t r(s_t, a_t)|s_0=s, a_0=a]
% \end{gathered}
% \end{equation}

%% file: framework.tex
In this section, we describe our proposed  approach to the taxi equilibrium policy learning problem. First, we introduce a mathematical model of ride pick-up probability incorporating the effect of traffic flow on a road network. We then introduce a method for estimating the equilibrium policy given the reward function $g(s,a)$. Finally, we introduce Spatial Equilibrium Inverse Reinforcement Learning (SEIRL), an algorithm for learning the cost function from the trajectories of multiple drivers. Our formulation is intended to model the passenger-seeking behavior on the streets, but it can be applied to the searching behavior for ride-hailing app requests simply by changing pick-up probability.

\begin{table}[t]
\caption{Notation.}
\vspace{-0.05in}
\label{table:params}
\centering
\begin{tabular}{|l|l|} \hline
{\bf Symbol} & {\bf Description} \\ \hline
$N$ & the number of vehicles \\ \hline
$S$ & state of the Markov game \\ \hline
% $\mathcal{A}$ & edges in the road network\\ \hline
$s^i,s$ & state (location on the road network) of agent $i$ \\ \hline
$a^i,a$ & action taken by the agent $i$ \\ \hline
% $c_a$ & cost of action $a$ \\ \hline
$\tau_a$ & travel time of action $a$ \\ \hline
% $r_s$ & average trip fare of passengers ridden at $s$ \\ \hline
$\mu^0_s$ & initial state distribution of state $s$ \\ \hline
$\mu_a(\mu_s)$ & visitation count of action $a$ (or state $s$)\\ \hline
$\rho_a$ & pick-up probability of action $a$ \\ \hline
$\lambda_s$ & arrival rate of potential passengers at $s$ \\ \hline
$\sigma_s$ & dropout rate of potential passengers at $s$ \\ \hline
$d_{s,s'}$ & destination distribution $s'$ of passengers at $s$ \\ \hline
$h_{s,s'}$ & ride trip time from $s$ to $s'$ \\ \hline
% $\mathcal{M}_x$ & $\{ \lambda_s, \sigma_s, d_{s,s'}, \tau_a, N\}$ \\ & parameters of taxi market model in context $x$ \\ \hline
% $\gamma$ & time discount rate \\ \hline
% $\beta$ & temperature of softmax policy \\ \hline
\end{tabular}
\end{table}

\begin{figure*}[t]
\centering
%\vspace{-0.1in}
\begin{subfigure}{0.26\textwidth}
\includegraphics[width=0.96\textwidth]{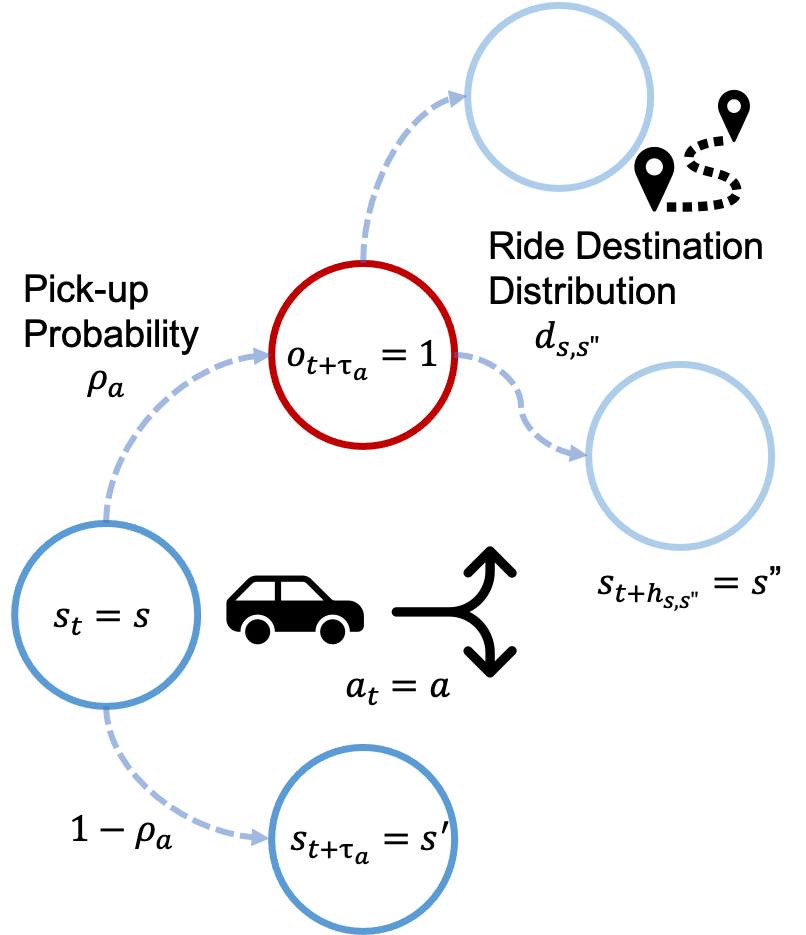}
\caption{MDP formulation}
\end{subfigure}
\begin{subfigure}{0.64\textwidth}
\includegraphics[width=0.96\textwidth]{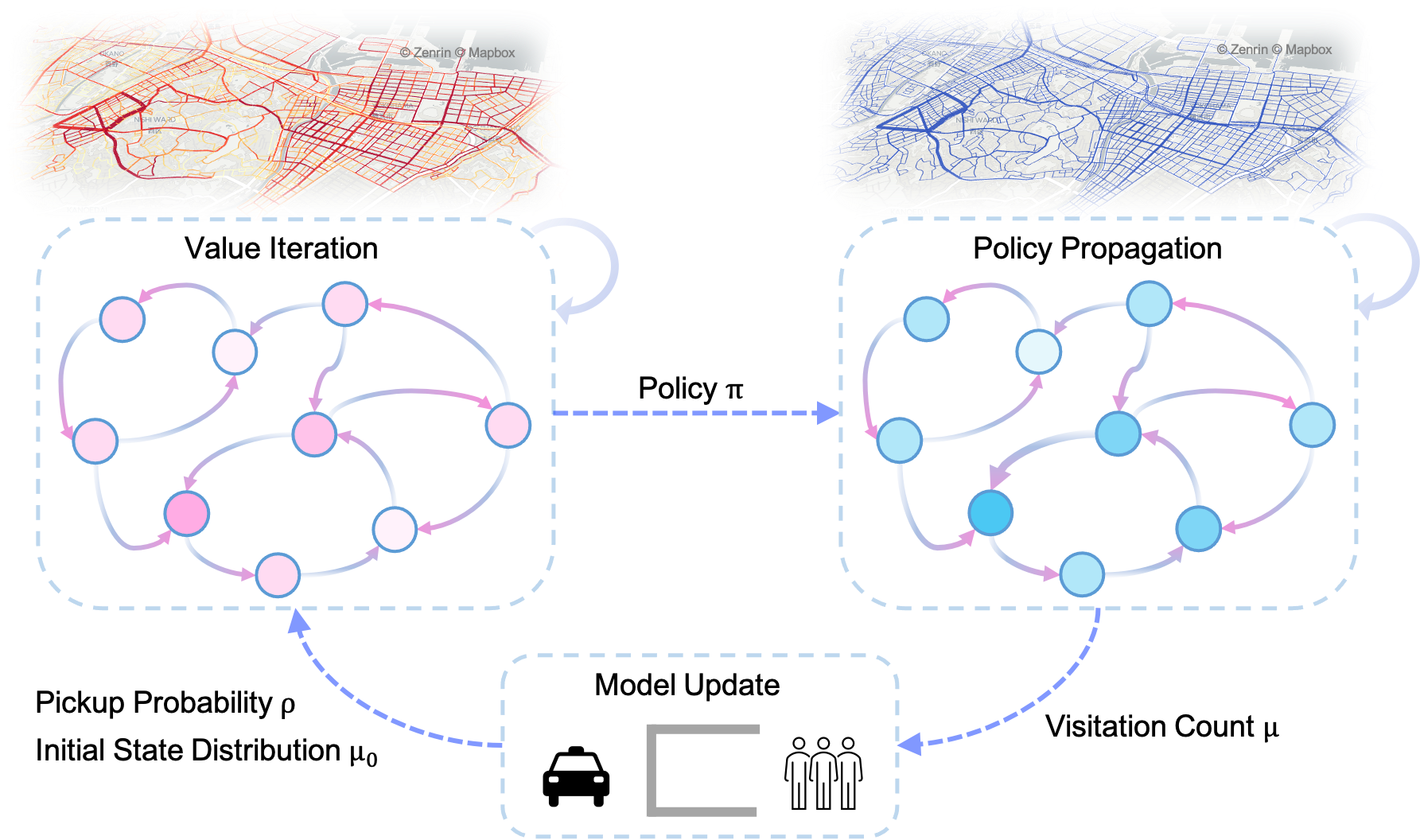}
\caption{Spatial Equilibrium Value Iteration}
% \label{fig:system}
\end{subfigure}

\caption{(a) Each state is represented as a node of the road network. (b) According to the current visitation count belief, the pick-up probability and the initial state distribution are updated first. Then, Value Iteration runs to find the optimal policy and Policy Propagation module probabilistically traverses the MDP given the current policy to compute the expected visitation count. The process is repeated until traffic flow converges.}
\label{fig:system}
\Description{}
\vspace{-0.1in}
\end{figure*}

\subsection{Pickup Probability Model}\label{sub:pp}
While pickups generated on individual road segments are observable, potential passengers who could not catch taxis on streets are not. In order to simulate a driver-passenger matching on a street, we need to estimate potential travel demand for each road segment from observable pickup data. We assume that the probability of the occurrence of a ride for each action should be expressed as a function of the potential demand for the location and the traffic flow rate, or state action visitation count, which is the frequency with which agents perform that action on average. We consider each road as an independent queue model such that a ride will occur if passengers are in the queue when the vehicle passes. Consider a situation where at time $t$, on road $s$, potential passengers occur at a arrival rate $\lambda_s^t$ and empty vehicles pass at a service rate $\mu_s^t$. There is an upper limit to passenger waiting time which depends on the time and location, and a passenger will disappear at a certain dropout rate $\sigma_s^t$. Assuming that the waiting queue length follows a Poisson distribution with an mean of $\lambda_s^t / (\mu_s^t + \sigma_s^t)$, the number of rides $m_s^t$ and the probability of ride occurrence $\rho_s^t$ are represented as:

\begin{equation}\label{eq:pp}
\begin{gathered}
\rho_s^t(\mu_s^t) = 1 - \exp \left( {\frac{\lambda_s^t}{\mu_s^t + \sigma_s^t}} \right) \\
m_s^t(\mu_s^t) = \mu_s^t \rho_s^t(\mu_s^t)
\end{gathered}
\end{equation}

where $\lambda_s^t$ and $\sigma_s^t$ are unknown parameters we want to estimate and $\mu$ is a given state action visitation count.
Given the observational daily draws of vehicle flow $\mu_{s,i}^t$ and the number of rides $m_{s,i}^t$, we can estimate $\lambda_s^t$ and $\sigma_s^t$ by the maximum likelihood estimation method using the following formula:

\begin{equation}\label{eq:mle}
\max_{\lambda, \sigma} \log \prod_i \binom{\mu_i}{m_i} \rho(\mu_i)^{m_i} (1 - \rho(\mu_i))^{\mu_i - m_i}.
\end{equation}
% = \max_{\lambda, \sigma} \sum_i (m_i \log \rho(\mu_i) + (\mu_i - m_i)\log (1 - \rho(\mu_i))).
Once we got parameters $\lambda_s^t$ and $\sigma_s^t$, we can compute pick-up probability $\rho_s^t$ for arbitrary visitation count $\mu$ by equation~\ref{eq:pp}.

\subsection{Equilibrium MDP formulation}
In general, although the dynamics of the environment, including all vehicle states, are often complex and difficult to model accurately, we can formulate the problem in a tractable way under mild assumptions that we will describe. 

Assuming that changes in environmental parameters, such as potential demand, are gradual with respect to the agent's discount rate $\gamma$, we no longer need to consider changes in environmental parameters over time and can assume that environmental parameters are fixed in the current context to determine the agent's policies. In this case, since environment is static, each agent acts according to a policy of equilibrium state reached if it has been in the current environment for an infinite amount of time. In other words, estimating policies for different time periods can be treated as different Markov games with different environmental parameters. For example, if we divide the time of day into 30-minute slots, agents act according to an equilibrium policy depending on demand and travel time during the 9:00-9:30 am period, but from 9:30-10:00 am, the environmental parameters change and the equilibrium policy changes accordingly. Thus, we do not include the time of day and season in the state, but call it a context when the transition probability changes, although the state action space is the same.

Since real taxi drivers do not have access to real-time supply-demand distribution, it is natural to expect that they are acting to maximize profitability based on the average supply-demand distribution obtained from their previous experience, not on current states of the environment. Thus, we further assume that, in equilibrium, pickup probability at time $t$ on the date $i$ can be replaced with expectation over daily draws $i$, i.e., $\rho_s^{t}(\mu_{s,i}^t) = \rho_s^{t}(\mu_s^t)$, which is uniquely determined regardless of the current state.
% and that the interaction of the entire agent only affects the dynamics through changes in the pickup probability of each road.
% Since real taxi drivers do not have access to real-time supply-demand distribution, it is natural to expect that they are acting to maximize profitability based on the average supply-demand distribution obtained from their previous experience, not on real time states of the environment.
As explained in \ref{sub:pp}, the pick-up probability for a given action depends on the demand and the state-action visitation count, meaning that the visitation count of current policy can completely determine the dynamics of the environment.

Given visitation count $\mu$, each agent policy becomes completely independent on other agents and we can reformulate the multi-agent policy learning problem corresponding to the result of repeatedly applying a stochastic entropy-regularized best response mechanism as the following single-agent forward RL problem:

\begin{equation}\label{eq:single-rl}
\max_{\pi} \mathbf{E}[\sum_{t}\gamma^{t}(g_{\theta}(s_{t},a_{t}) - \log \pi(a_t|s_t))|s_t, a_t,\pi,\mu^{\pi}].
\end{equation}

% Suppose we know visitation count at the equilibrium $\mu$.Given environmental parameters $\mathcal{M}_x = \{\lambda_s, \sigma_s, d_{s,s'}, \tau_a, N, G(\mathcal{V}, \mathcal{E}) \}$ and 
Suppose we know the reward function $g_{\theta}(s,a)$. Then, the optimal policy can be computed by iterating following Bellman backups in a single-agent MDP:

\begin{equation}\label{eq:vi}
\begin{aligned}
Q(s,a) = &
g_{\theta}(s,a) + \rho_a(\mu_s) \sum_{s''} d_{s,s''} \gamma ^{h_{s,s''}} V(s'') \\ & + (1 - \rho_a(\mu_s)) \gamma ^{\tau_a} V(s'), \\
V(s) = & \text{soft}\max_{a} Q(s, a),
\end{aligned}
\end{equation}

where we assume the reward function can be represented as $g_{\theta}(s,a) = \rho_a(\mu_s)\bar{w_s} - \theta \cdot f_{s,a}$, which is expected pick-up bonus (pick-up probability $\rho_a(\mu_s)$ multiplied by expected fare revenue $\bar{w_s} = \sum_{s'} d_{s,s'} w_{s,s'}$) minus linear action cost dependent on features $f_{s,a}$ associated with $s,a$. Since taxi drivers in many countries, including Japan, work on a commission basis, it is natural to think of them as basically acting to minimize idle time and maximize their income. However, many other factors, including familiarity with the area and the level of danger on that road, can be considered as a cost for drivers.

% Note that $w_{s,s'}$ and $h_{s,s'}$ are determined depending on the road network distance between $s$ and $s'$.
% where reward function $g_{\theta}(s,a) = 

The shared drivers policy is given by
\begin{equation}\label{eq:policy}
\pi(a|s) = \exp (Q(s,a) - V(s)).
\end{equation}

The visitation count $\mu^{\pi}$ corresponding to traffic flow rate can be obtained by propagating state distributions of $N$ agents by executing policy $\pi$ starting from passenger drop-off distribution $d_s$ as initial state distribution. Since we assume that the total number of vehicles $N$ is constant and the average total number of empty vehicles can be calculated as $\mu \cdot \tau$ by Little's law, the average drop-off rate, i.e. initial state distribution, is approximated by $\mu^0_s = d_s (N - \mu \cdot \tau) / h$, where $h$ is average ride trip time.
Given initial state distribution $\mu^0_s$, the visitation count of policy $\pi$ satisfies
\begin{equation}\label{eq:visitation}
\mu_s = \mu^0_s + \sum_{s',a} (1 - \rho_a)\pi(a|s')\mu_{s'}.
\end{equation}

We can compute the policy and visitation count at the equilibrium which holds Equations (\ref{eq:vi}), (\ref{eq:policy}), (\ref{eq:visitation}) by iterating Value Iteration and policy propagation, presented in Algorithm~\ref{alg:sevi}.
First, at the beginning of each iteration, we update the pick-up probability and the initial state distribution according to the previous visitation count, and then we iterate Bellman backups according to Equation \ref{eq:vi} to find the optimal policy. Next, we compute the expected visitation count by probabilistically traversing the MDP given the current policy, starting from the initial state distribution. The weighted sum of the previous iteration is used in order to suppress the abrupt changes and to stabilize the algorithm. The process is repeated until mismatch distance ratio between $\mu_k$ and $\mu_{k-1}$ falls below the threshold, i.e., $\mathcal{M}(\mu_k,\mu_{k-1}) < \epsilon$ where $\mathcal{M}$ is given by Equation \ref{eq:mdr}.

\begin{algorithm}[h]
\begin{algorithmic}[1]
\caption{Spatial Equilibrium Value Iteration}\label{alg:sevi}
\State Input reward function $g_{\theta}$
\State Set counter $k = 0$
\State Initialize visitation count belief $\mu_k$
\State Fix parameter $\alpha, \beta, \gamma$
\Repeat
\State Update pick-up probability $p(\mu_k)$ per Equation \ref{eq:pp}
\State Update initial states. $\mu^0 = \frac{d}{h}(N - \mu_k \cdot \tau)$
\State Compute $V(s), Q(s,a)$ by iterating Equation \ref{eq:vi}
\State Compute policy $\pi_k(a|s) \gets \exp(\frac{1}{\beta}(Q(s,a) - V(s)))$
\State Compute $\mu_s$ by propagating policy $\pi_k$
\State Update visitation count belief $\mu_k \gets (1 - \alpha) \mu_{k-1} + \alpha \mu $
\State $k \gets k + 1$
\Until{$ \mathcal{M}(\mu_k, \mu_{k-1}) < \epsilon$}
\end{algorithmic}
\end{algorithm}

\subsection{Equilibrium IRL}
Given the expert's visitation count $\mu^E$, our goal is to estimate the parameters $\theta$ of the reward function $g_{\theta}(s,a)$ such that the visitation count in equilibrium is consistent with $\mu^E$.
Similar to Equation (\ref{eq:single-rl}), we can simplify multi-agent IRL problem by decomposing likelihood function in Equation (\ref{eq:ma-irl}) into probability of individual agents' trajectories.

\begin{equation}
\begin{gathered}\label{eq:mg-irl}
L(\theta) = E_{\bm{\zeta} \sim D} [\log \bm{p}_{\theta}(\bm{\zeta})] = E_{\zeta^i \sim D} [\log p_{\theta}(\zeta^i)] \\
p_{\theta}(\zeta) = \mu^0(s_0) \prod_{t=0}^{T}p(s_{t+1}|s_t,a_t;\mu)\pi_{\theta}(a_t|s_t;\mu)) 
\end{gathered}
\end{equation}

% We assume that they are maximizing overall rewards that include not only simple profitability, but also drivers' area preferences, road safety, passenger type, etc.
\begin{theorem}
Given reward function $g_{\theta}$, let $\mu^E_{s,a,s'} = \mu^E_{a} p(s'|s,a;\mu^E)$ denote visitation frequencies of state $s$, action $a$ and next state $s'$ in the expert trajectories $\mathcal{D}$, let $\mu^{\pi}_{s,a,s'} = \mu^{\pi}_{a} p(s'|s,a;\mu^{\pi})$ denote visitation frequencies of policy $\pi$. Then the gradient of the likelihood in Equation (\ref{eq:mg-irl}) is given by:
\begin{equation}\label{eq:true-grad}
\begin{aligned}
\frac{\partial L}{\partial \theta} &= \sum_{s,a,s'} (\mu^E_{s,a,s'} - \mu^{\pi}_{s,a,s'}) \frac{\partial}{\partial \theta} \left( g_{\theta} + \log p^{\pi}(s'|s,a) \right) \\
&= \sum_{s,a} (\mu^E_a - \mu^{\pi}_a) \frac{\partial g_{\theta}}{\partial \theta} + \mu^E_a \left( \frac{\rho_a(\mu^E_a) - \rho_a(\mu^{\pi}_a)}{1 - \rho_a(\mu^{\pi}_a)} \right) \frac{\partial}{\partial \theta} \log \rho_a(\mu^{\pi}_a)
\end{aligned}
\end{equation}
Proof. See Appendix \ref{sub:appendix-mle}
\end{theorem}
Unlike the case of a normal single agent, the transition probability also depends on $\theta$ through $\mu$ and computing the gradient of the second term is not tractable. We further assume $\rho << 1$ and substitute the Taylor expansion of Equation \ref{eq:pp}:

\begin{equation}\label{eq:grad}
\begin{aligned}
\frac{\partial L}{\partial \theta} &\approx \sum_{s,a} (\mu^E_a - \mu^{\pi}_a) \left( \frac{\partial g_{\theta}}{\partial \theta} + \rho_a(\mu^E_a) \left( \frac{\mu^E_a}{\mu^{\pi}_a+\sigma_a - \lambda_a} \right) \frac{\partial}{\partial \theta} \log \rho_a(\mu^{\pi}_a) \right) \\
&\approx \sum_{s,a} (\mu^E_a - \mu^{\pi}_a) \frac{\partial g_{\theta}}{\partial \theta}
\end{aligned}
\end{equation}

As can be seen, we have shown that the gradient of the likelihood of multi-agent trajectories can approximate the gradient of MaxEnt IRL, under the assumption $\rho << 1$, which is satisfied in most cases of ride-hailing services. In addition, as $\rho_a$ depends on $\theta$ through $\mu$ and computing the gradient is intractable, we drop the pick-up bonus term $\rho_a(\mu_s) \bar{w_s}$ from reward function $g_{\theta}(s,a)$. We empirically confirmed that, without pick-up bonus term, reward learning becomes more stable and robust.
The entire reward learning process is shown in Algorithm~\ref{alg:seirl}, which iterates through the estimation of the equilibrium policy and the visitation count in Algorithm~\ref{alg:sevi} and reward parameter updates with exponentiated gradient descent. In order to prevent abrupt change in weight $\theta$, we divide gradient by $\sqrt{|\Delta|}$.

\begin{algorithm}[h]
\begin{algorithmic}[1]
\caption{Spatial Equilibrium IRL}\label{alg:seirl}
\State Input expert demonstrations $\mathcal{D}$
\State Initialize weight vector $\theta$
\For{$t\gets 1, T$}
\State Sample expert visitation count $\mu^E$ from $\mathcal{D}$
\State Compute policy $\pi$ and visitation count $\mu^{\pi}$ by Algorithm 1.
\State Compute gradient $\Delta=(\mu^E - \mu^{\pi})\cdot f$
\State Update $\theta \gets \theta \exp(-\eta \Delta / \sqrt{|\Delta|})$
\State Update reward $g \gets -\theta \cdot f$
\EndFor
\end{algorithmic}
\end{algorithm}

%% file: experiments.tex
We showcase the effectiveness of the proposed framework for addressing changes in dynamics and data losses in series of simulated experiments using real taxi trajectory data.

\subsection{Data Preparation}

\begin{figure}[h]
  \includegraphics[width=0.4\textwidth]{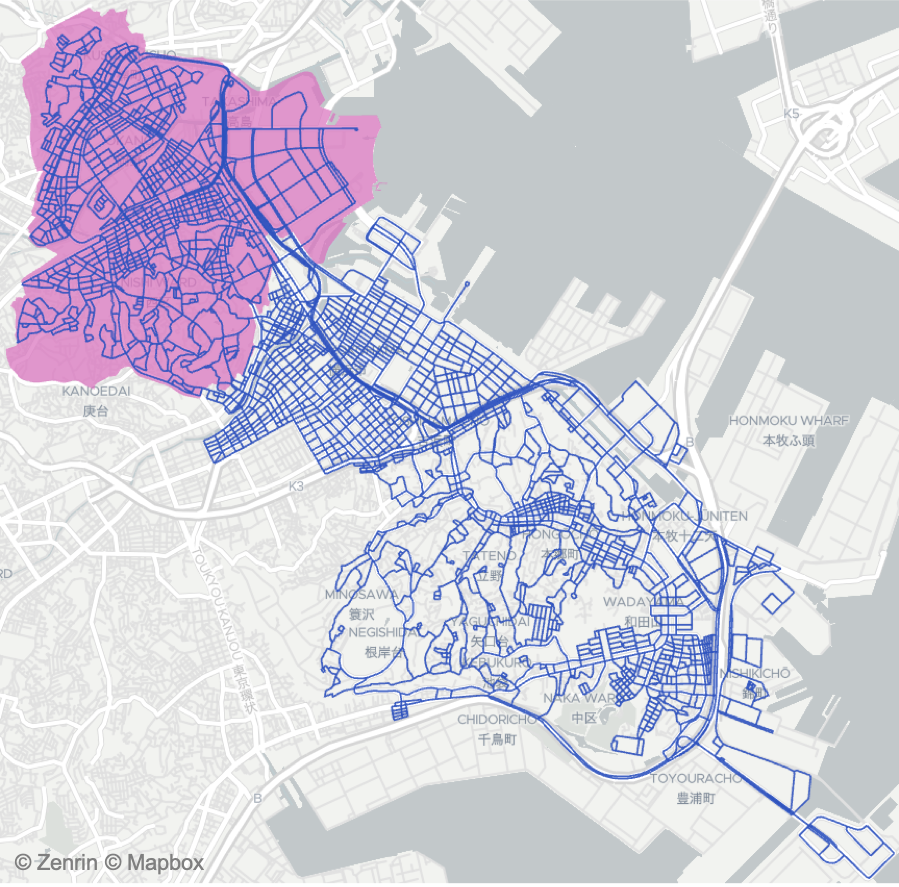}
  \caption{Road Network in Yokohama city}
  \Description{Width of lines describe visitation count of vacant taxi}
  \label{fig:area}
\end{figure}

For data analytics and performance evaluation, we used taxi trajectories, collected in the city of Yokohama, Kanagawa, Japan during the period between June 2019 and June 2020 sampled at a period of 3-10 seconds. The original raw dataset contains the driver id, trip id, latitude, longitude, and timestamp of the empty vehicle. The first record of each trip id stands for drop off point or market entry point and the last record stands for a pick up point. As a pre-processing, the GPS information was linked to the roads by map-matching, and the routes between the roads were interpolated with the shortest routes to enable aggregation of the data per road, which resulted in low loss ratio. The road network in the region contained 10765 road segments (nodes) and 18649 transitions (edges).

The road travel time $\tau_a$ was averaged over the time spent on the same road, and any missing data were filled in with the average for each road class.

The behavioral model was evaluated for the period 7:00-22:00 Monday-Thursday. The late night to early morning period was omitted from the evaluation period because there are fewer passengers on streets. For the evaluation, the data set was divided into the following four periods.

\begin{itemize}
    \item Profile: 12 weeks starting from 2019-04-01
    \item Train: 12 weeks starting from 2019-07-01
    \item 19Dec: 8 weeks starting from 2019-12-12
    \item 20Apr: 4 weeks starting from 2020-04-01
\end{itemize}

In each dataset, we treated each 30-minute period as a different context, and in each context we aggregated the number of rides $m_s^t$, the number of empty vehicles passing through $\mu_s^t$ and the number of vehicles in initial condition $\mu_s^0$ on a road-by-road basis. To fit the pick-up probability model $\lambda_s^t, \sigma_s^t$ in Equation (\ref{eq:pp}), we used daily draws of $\mu_{s,i}^t, m_{s,i}^t$ in Profile dataset.

\begin{figure}[h]
  \includegraphics[width=0.35\textwidth]{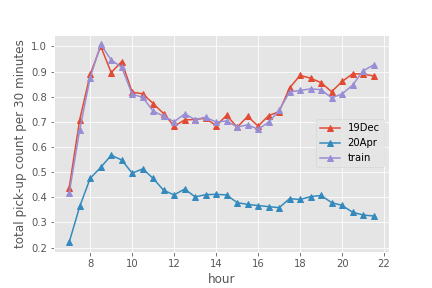}
  \caption{Average pick-up count per 30 minutes for each dataset. Count is normalized by max value of Train dataset.}
  \label{fig:pickup_hhb}
\end{figure}

The number of half-hourly rides in each dataset is shown in Figure~\ref{fig:pickup_hhb}. While Train dataset was collected in summer 2019, the period of 19Dec is colder holiday season of winter 2019, indicating that 19Dec has a different context than Train dataset at the same hour of day.
For 20Apr, the environment is significantly different due to the sharp drop in demand due to COVID-19.  By comparing the accuracy in these datasets, we assessed the robustness of the learned policies to the impact of changing dynamics on seasonal variations.

\subsection{Implementation Details}
For reward features, we use map attributes including travel time, road class and type of turn, in addition to the statistical features extracted from the Profile dataset, such as how frequent the road was traversed. Additionally, to incorporate unobservable fixed effects of certain actions, we include unique features associated with each actions. This allows the cost of each action to vary independently.
Also, L2 regularization is added to the gradient for preventing overfitting. Training iteration is 100 and learning rate is 1.0,  multiplied by 0.98 for each iteration to stabilize training. We partitioned time by 3 hours and learned different cost weights for each time zone: 7:00-10:00, 10:00-13:00, 13:00-16:00, 16:00-19:00, 19:00-22:00.

% ?Reward function $g_{\theta}$ can be any function approximator including neural networks, but feature vector have to be independent on $\mu$, which depends on $\theta$ through $\pi_{\theta}$. So we does not include pick-up probability into features.

\subsection{Evaluation Methods}
We compare our apporach with the following baselines.

\noindent \textbf{Opt}: We let cost to be travel time $g(s,a) = -\tau_a$ and obtain the policy by Value Iteration when initial visitaion count belief is the one in the training dataset.

\noindent \textbf{SE-Opt}
We let cost to be travel time $g(s,a) = -\tau_a$ and obtained equilibrium policy by Algorithm~\ref{alg:sevi}.

\noindent \textbf{Tr-Expert}
Expert policy is calculated from the statistics of the training dataset period $\pi(a|s) = \sum_i 1(a_i=a) / \sum_i 1(s_i=s)$.

The Mismatch Distance Ratio (MDR) was used as an accuracy measure of the proximity of the obtained policies to the expert policies, which is given by

\begin{equation}\label{eq:mdr}
\mathcal{M}(\mu^E, \mu^{\pi}) =  \frac{\sum_{a} |\mu^E_a - \mu^{\pi}_a| l_a }{\sum_{a} \mu^E_a l_a}
\end{equation}
where $l_a$ is the length of the road.

The evaluation of SEIRL was performed as follows. First, we estimate the parameters of the cost function to fit the expert visitation count of Train dataset. When evaluating 19Dec or 20Apr dataset, using the environment parameters of each context (every 30 minutes between 7:00 and 22:00) and the learned cost function associated with the context, Algorithm~\ref{alg:sevi} is used to estimate the equilibrium policy. We then calculate the equilibrium visitation count $\mu^{\pi}$ by repeating the policy propagation and compare it with $\mu^E$ by MDR.

%% file: results.tex
\begin{table*}[t]
\caption{Summary of imitation performance measured by mismatch distance ratio}
\vspace{-0.1in}
\label{table:results}
\centering
\begin{tabular}{lrrrrrrrr}
\toprule
{} & \multicolumn{2}{l}{Normal} & \multicolumn{2}{l}{Disabled} & \multicolumn{2}{l}{Defect5\%} & \multicolumn{2}{l}{Defect10\%} \\
{} &  19Dec &  20Apr &    19Dec &  20Apr &    19Dec &  20Apr &     19Dec &  20Apr \\
policy    &        &        &          &        &          &        &           &        \\
\midrule
Opt       &  2.867 &  2.381 &    2.780 &  2.321 &    2.867 &  2.381 &     2.867 &  2.381 \\
SE-Opt    &  0.943 &  1.076 &    0.923 &  1.078 &    0.943 &  1.076 &     0.943 &  1.076 \\
SEIRL     & \bf 0.223 &  \bf 0.328 &    \bf 0.291 &  \bf 0.391 &    \bf 0.363 &  \bf 0.485 &     \bf 0.411 &  \bf 0.543 \\
Tr-Expert &  0.230 &  0.349 &    0.726 &  0.813 &    0.437 &  0.553 &     0.542 &  0.642 \\ \hline
Expert    &  0.099 &  0.123 &    0.128 &  0.160 &    0.099 &  0.123 &     0.099 &  0.123 \\
\bottomrule
\end{tabular}
\vspace{-0.1in}
\end{table*}

% \subsection{Trend Change}

% This section describes the results of experiments.
% First, We assessed the robustness of the taxi market to seasonal factors.

% The number of half-hourly rides in each dataset is shown in Figure~\ref{fig:pickup_hhb}. While Train dataset was collected in summer 2019, the period of 19Dec is colder holiday season of winter 2019, indicating that 19Dec has a different context than Train dataset at the same hour of day.
% For 20Apr, the environment is significantly different due to the sharp drop in demand due to COVID-19.  By comparing the accuracy in these datasets, we assessed the robustness of the learned policies to the impact of changing dynamics on seasonal variations.

% \begin{figure}[h]
%   \includegraphics[width=0.35\textwidth]{pickup_hhb.png}
%   \caption{Average pick-up count per 30 minutes for each dataset. Count is normalized by max value of Train dataset.}
%   \label{fig:pickup_hhb}
% \end{figure}

Using the reward function learned from Train dataset, we evaluated imitation performance of each policy for test datasets (19Dec and 20Apr). The MDR for each policy is shown in the Normal column of Table \ref{table:results}. SEIRL scored the best for both environments, with an error of about 33\% during COVID-19 pandemic. There is a large gap between SE-Opt and SEIRL, which can be interpreted as drivers minimizing a more complex reward function rather than simply taking the shortest possible route to potential customers. The fact that SE-Opt is significantly better than Opt indicate that computing the policy in equilibrium has a significant impact on improving imitation accuracy. 

\begin{figure}[h]
\centering
\begin{subfigure}{0.23\textwidth}
\includegraphics[width=0.96\textwidth]{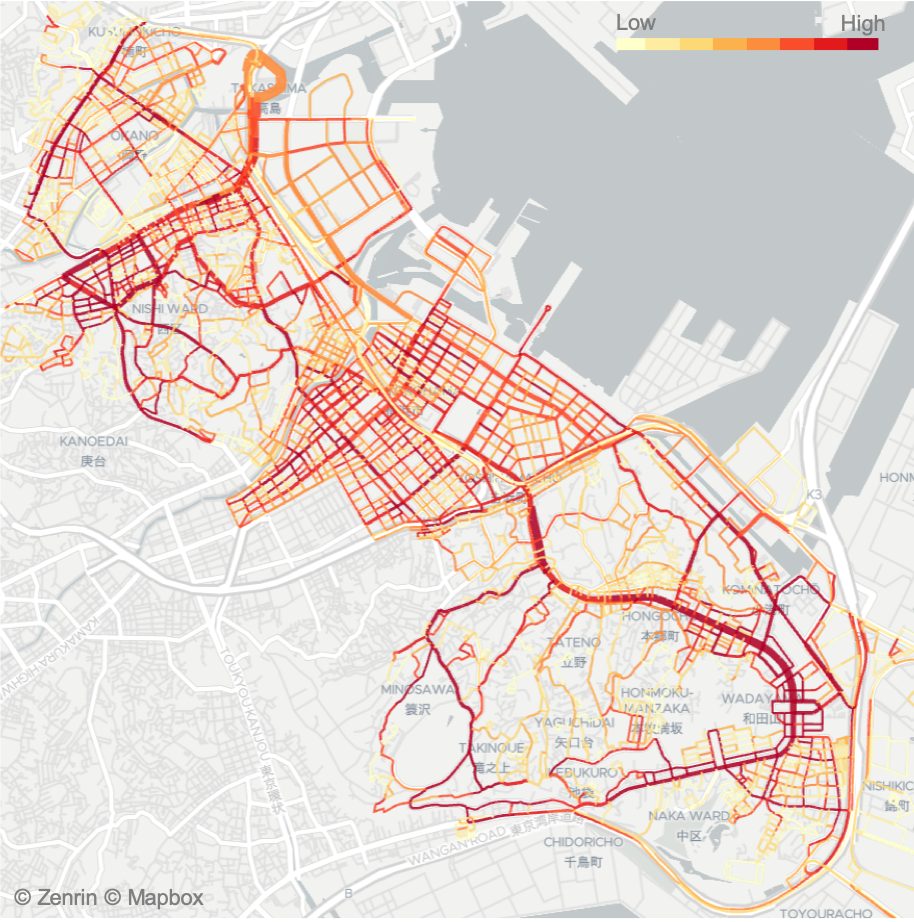}
\caption{8:00-8:30}
\end{subfigure}
\begin{subfigure}{0.23\textwidth}
\includegraphics[width=0.96\textwidth]{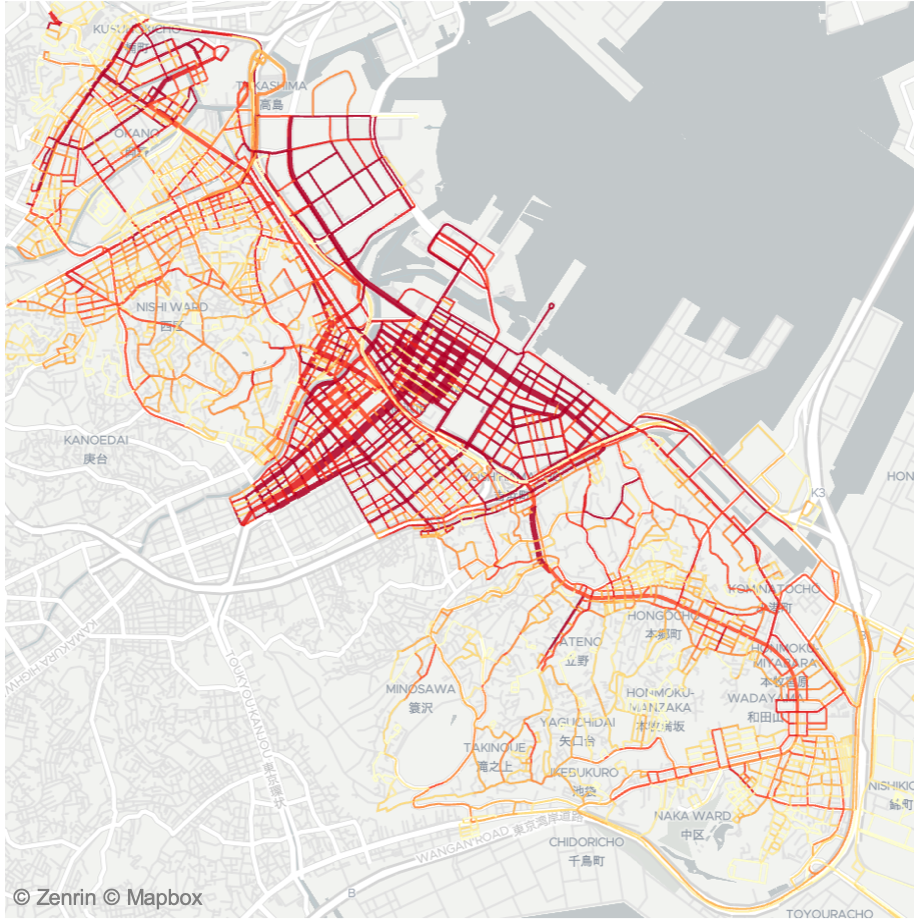}
\caption{21:30-22:00}
\end{subfigure}
\caption{Visualizations of state value at different times computed by Algorithm \ref{alg:sevi}}
\label{fig:value}
\end{figure}

Figure \ref{fig:value} visualizes examples of the estimated state value during morning and evening. The clear difference in two geospatial distributions shows that drivers use significantly different policies depending on the context.

The computation time required to update the policy is polynomial to the network size, but not dependent on the number of agents. Depending on the hardware environment, with the problem size of Yokohama city, the policy can be updated in a few seconds even on a single CPU, which is sufficient for real-time computing applications.

\subsection{Disabled Area}
In order to evaluate the impact of a larger change in dynamics, we next conducted an experiment in which demand in Nishi-ku region (the pink area in Figure~\ref{fig:area}) is artificially eliminated. Since the spatial distribution of potential demand changes significantly, the actual policy is expected to be very different from the policy in the training enviroment. The ground truth visitation count was created by removing trips that occurred in Nishi-ku region over the same period for the datasets. The result is shown in the Disabled column in Table 2, where SEIRL is the best, significantly ahead of Tr-Expert. These results demonstrate that SEIRL has a high generalization performance.

\subsection{Data Loss}
\begin{figure}[h]
  \includegraphics[width=0.5\textwidth]{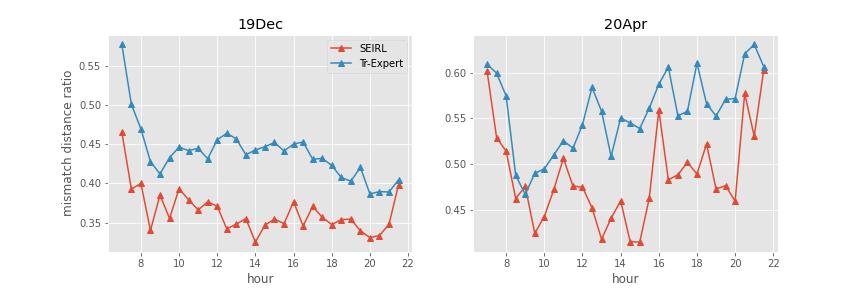}
  \caption{Mismatch distance ratio from 7:00 to 22:00}
  \Description{}
  \label{fig:def05_hhb}
\end{figure}

Furthermore, since real-life vehicle logs may be missing for various reasons, we evaluated the impact of missing data. The dataset used so far composed of trajectories interpolated by connecting GPS data with the shortest path in order to eliminate the influence of data noise. For actual applications, it is desirable to have good generalization performance even when training on a missing dataset. Therefore, we randomly sampled the nodes (states) of the road network and created the defected training dataset and features with all the data of those states missing, and performed the same evaluation. No specific changes were made to the evaluation dataset. The experiments were conducted for two conditions with a missing percentage of 5\% and 10\%, and the SEIRL had the lowest MDR in both conditions. In addition, the ratio of MDR deterioration to Normal at 19Dec/20Apr in the 5\% deficiency condition was 1.89/1.59 for Tr-Expert compared to 1.63/1.48 for SEIRL. In the 10\% deficiency condition, the accuracy deterioration was 2.35/1.84 for Tr-Expert and 1.84/1.65 for SEIRL, which demonstrates that SEIRL is more robust to missing data. Half-hourly MDRs at 5\% loss are shown in Figure~\ref{fig:def05_hhb}. We confirmed that SEIRL outperformed Tr-Expert at almost all time points.

\subsection{Counterfactual Simulation}
\begin{figure}[h]
  \includegraphics[width=0.48\textwidth]{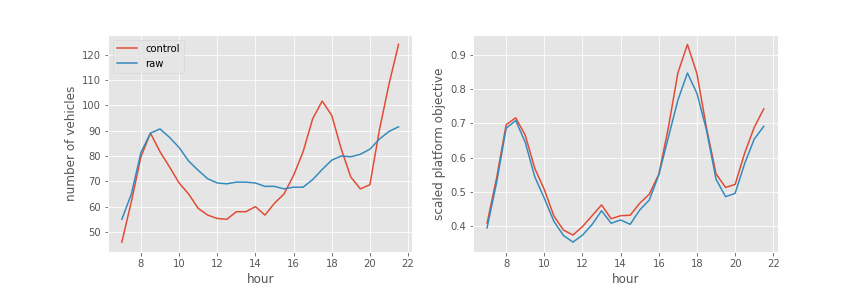}
  \caption{Comparison of actual values and control to optimal supply in the simulation}
  \label{fig:cf}
\end{figure}

Finally, as an example of counterfactual simulation using the learned driver behavior model, we demonstrate the optimization of the quantity of supply in a ride-hailing service. The market supply is often determined by the drivers' decision to enter the market and is not directly controlled. However, it should be possible to adjust the number of drivers entering the market indirectly to some extent by providing some kind of information or incentive to the drivers. The learned driver behavior model can be used to simulate at an arbitrary number of vehicles $N$, thus allowing us to find the optimal amount of supply. Now, suppose that the platform objective is represented by the following:

\begin{equation}
\max_N \sum_{s,a} (\rho_a(\mu_a^N) \mu_a^N \bar{w_s} - b \mu_a^N \tau_a)
\end{equation}
Here, the first term corresponds to the average revenue on the road $s$, i.e., the number of rides multiplied by the average fare per ride at $s$. The second term represents total cost including fuel and personnel fee, which is assumed to be proportional to driving time. $b$ is proportionality constant. $\mu_a^N$ is state-action visitation count produced by executing the learned policy in $N$-agents simulation.
We varied the number of vehicles per half hour from the current rate of 0.5 to 1.5, and determined the number of vehicles at each time slot with the largest platform objective. The optimized number of vehicles and the objective for $b=1.5$ are shown in Figure \ref{fig:cf}. It can be seen that the number of vehicles is lower than the current supply from the end of the morning peak to around 16:00, while it should be increased after 16:00~18:00 and 20:30.

\begin{figure*}[t]
\centering
%\vspace{-0.1in}
\begin{subfigure}{0.32\textwidth}
\includegraphics[width=0.96\textwidth]{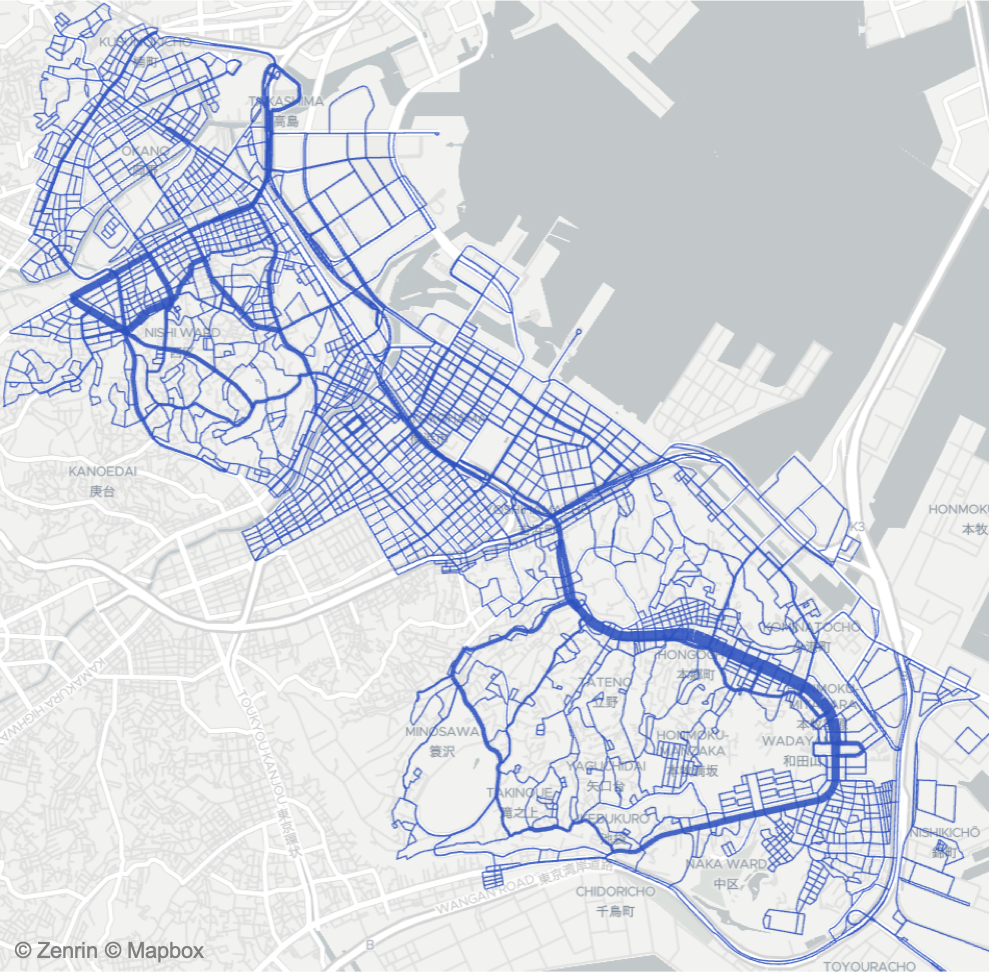}
\caption{Normal-20Apr: Expert}
\end{subfigure}
\begin{subfigure}{0.32\textwidth}
\includegraphics[width=0.96\textwidth]{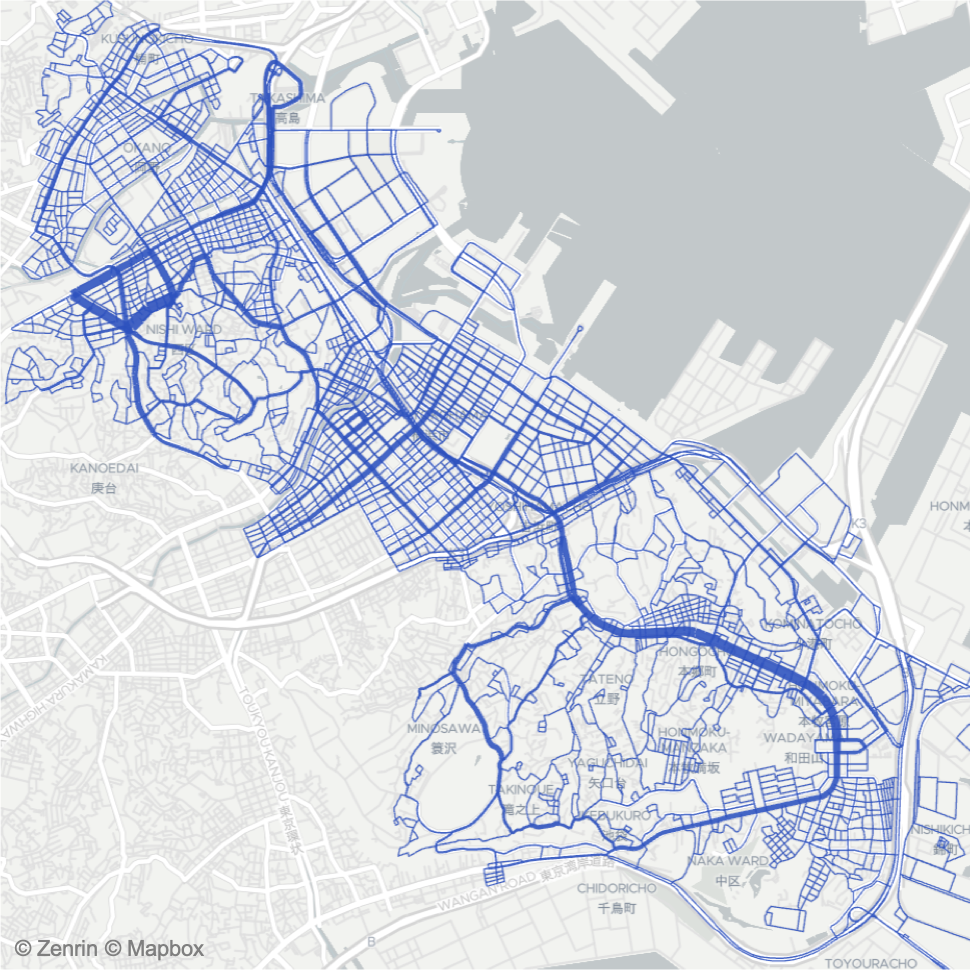}
\caption{Normal-20Apr: SEIRL}
\end{subfigure}
\begin{subfigure}{0.32\textwidth}
\includegraphics[width=0.96\textwidth]{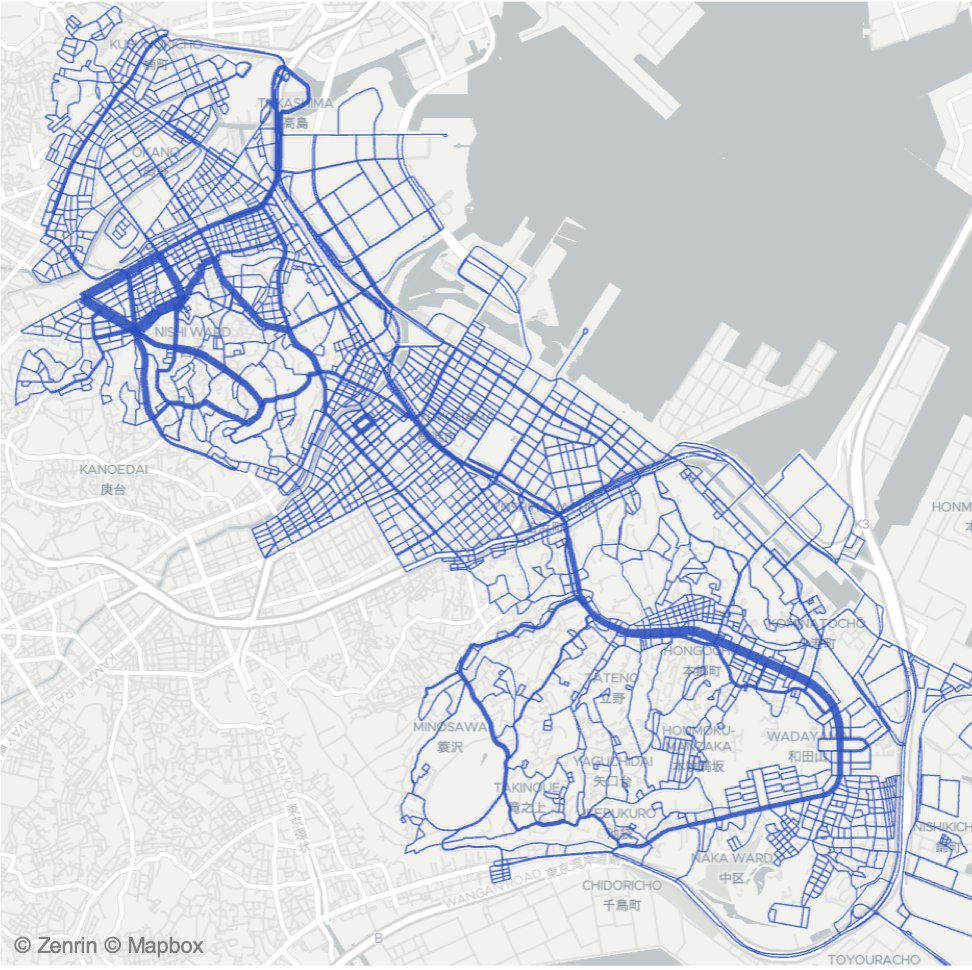}
\caption{Normal-20Apr: Tr-Expert}

\end{subfigure}
\begin{subfigure}{0.32\textwidth}
\includegraphics[width=0.96\textwidth]{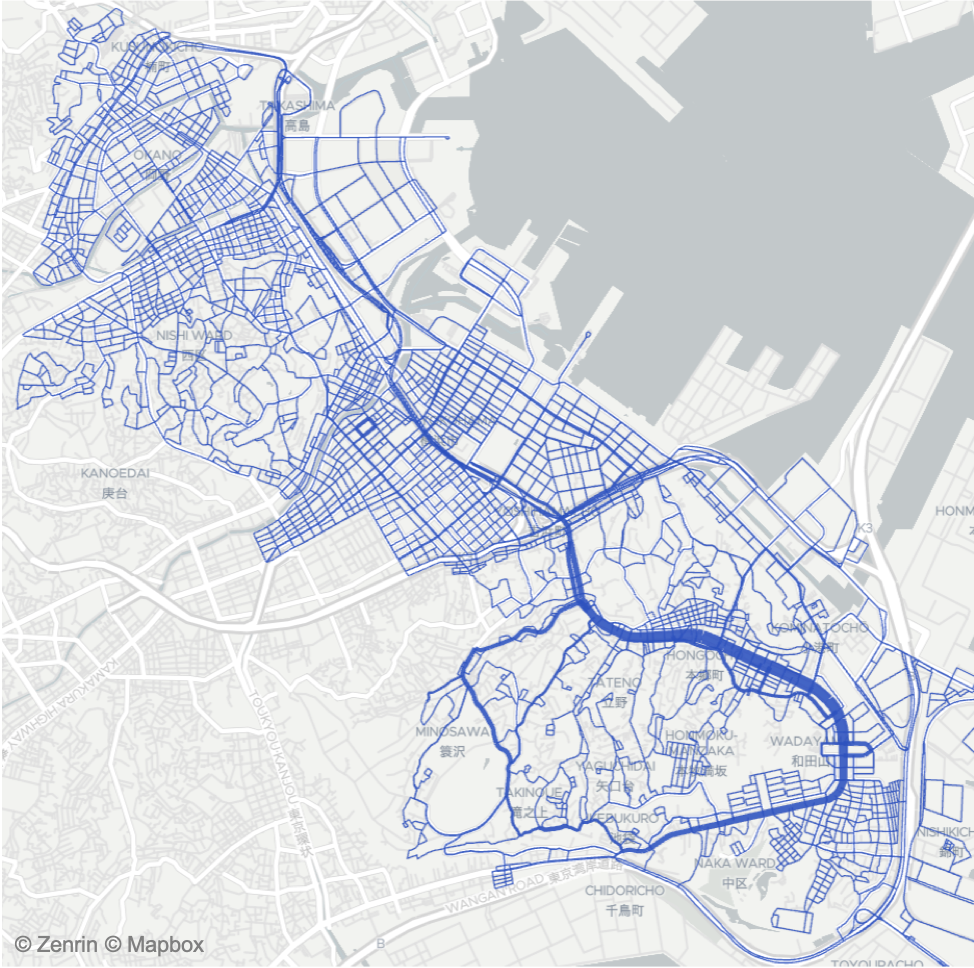}
\caption{Disabled-20Apr: Expert}
\end{subfigure}
\begin{subfigure}{0.32\textwidth}
\includegraphics[width=0.96\textwidth]{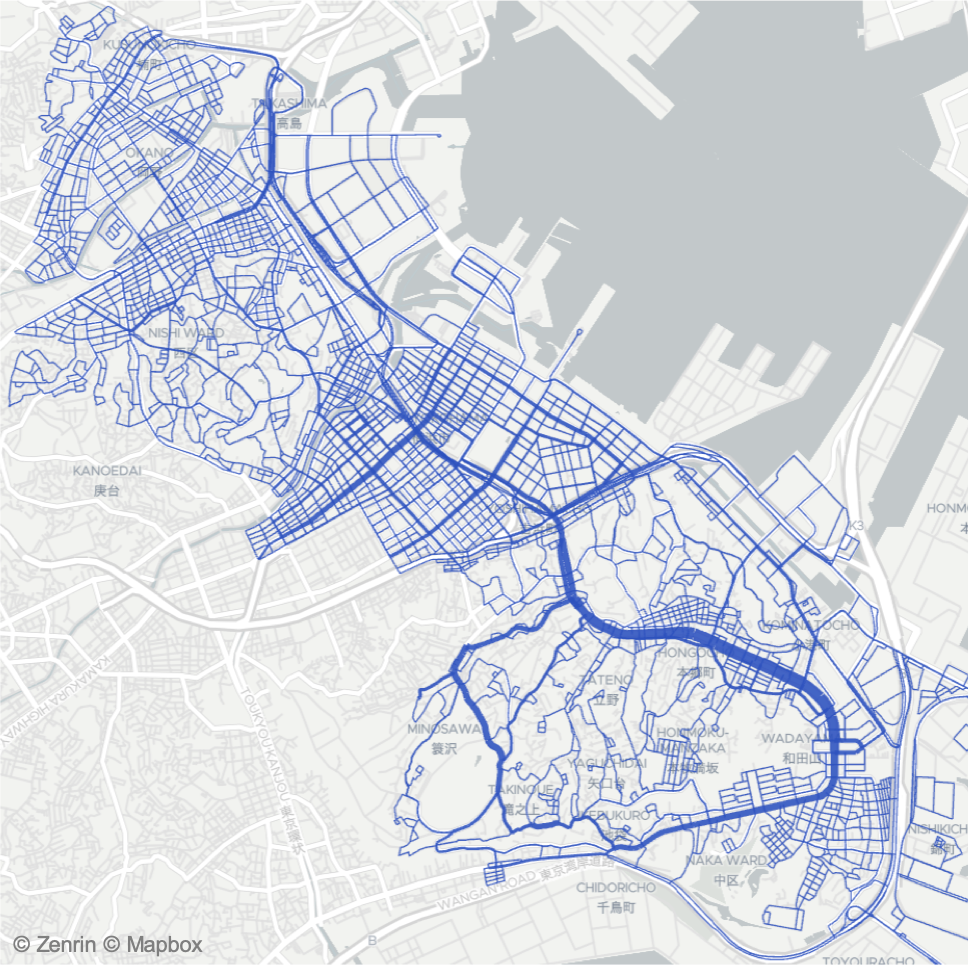}
\caption{Disabled-20Apr: SEIRL}
\end{subfigure}
\begin{subfigure}{0.32\textwidth}
\includegraphics[width=0.96\textwidth]{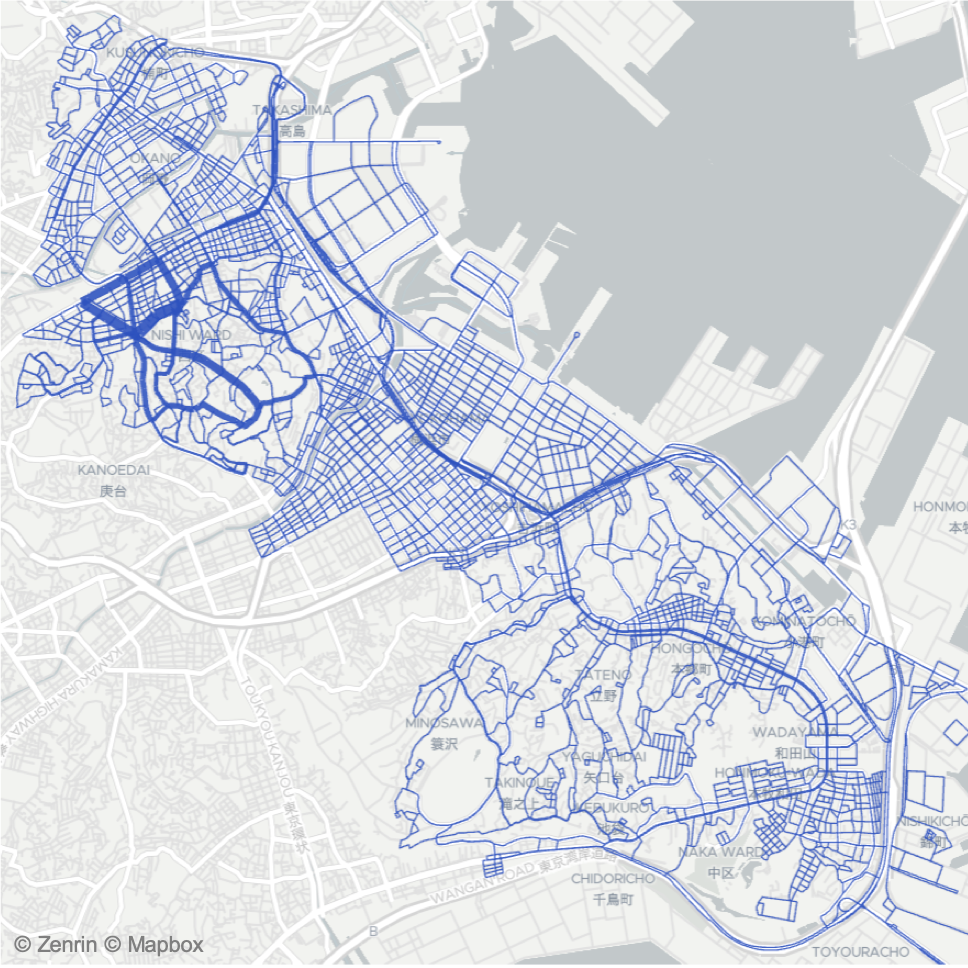}
\caption{Disabled-20Apr: Tr-Expert}
\end{subfigure}

\caption{Comparison of equilibrium visitation count during 8:00-8:30 executing the different policies.}
\vspace{-0.1in}
\end{figure*}

%% file: conclusions.tex
This paper proposed the first approach to multi-agent behavioral modeling and reward learning in equilibrium on a road network for a ride-hailing service. We focused on a ride-hailing vehicle network where each driver is uncooperatively searching for passengers on streets. Our proposed MDP formulation enables efficient and scalable simulation by computing the optimal policy in equilibrium shared between agents. We developed a reward learning framework integrated MaxEnt IRL and demonstrated that our method is able to recover policies that are robust to significant variation in dynamics.

Although our proposed method assumes all agents are homogeneous, in reality, driver behavior patterns can vary due to several factors. A possible extension could include learning routing patterns of heterogeneous agents considering multiple garage locations (cab offices). One may also be interested in equilibrium modeling that takes into account the effect of travel time due to congestion such as traffic jams, in addition to matching with passengers.

%% file: appendix.tex
\subsection{Maximum Likelihood Estimation for SEIRL}\label{sub:appendix-mle}
In this section, we include derivation of Equation (\ref{eq:true-grad}) and (\ref{eq:grad}).
From Equation (\ref{eq:mg-irl}), the likelihood function is given by:
\begin{equation}
\begin{aligned}
L &= \mathbf{E}_{\zeta \sim \mathcal{D}} \left[ \log \mu^0(s_0) \prod_{t=0}^{T}p(s_{t+1}|s_t,a_t;\mu^{\pi})\exp(g_{\theta}(s_t,a_t)) \right] \\ & - |\mathcal{D}| \log Z \\
Z &= \sum_{\zeta} \left[ \mu^0(s_0) \prod_{t=0}^{T}p(s_{t+1}|s_t,a_t;\mu^{\pi})\exp(g_{\theta}(s_t,a_t)) \right]
\end{aligned}
\end{equation}

Take derivatives of $L(\theta)$:
\begin{equation}
\begin{aligned}
\frac{\partial L}{\partial \theta} &= \mathbf{E}_{\zeta \sim \mathcal{D}} \left[ \sum_{t=0}^{T} \frac{\partial}{\partial \theta} g'_{\theta}(s_t,a_t, s_{t+1}) \right] \\
&- \frac{|\mathcal{D}|}{Z} \frac{\partial}{\partial \theta} \sum_{\zeta} \left[ \mu^0(s_0) \exp(\sum_{t=0}^{T}g'_{\theta}(s_t,a_t, s_{t+1})) \right] \\
&= \sum_{s,a,s'} \left[ \mu^E_{s,a,s'} \frac{\partial}{\partial \theta} g'_{\theta}(s,a,s') \right] \\
&- |\mathcal{D}| \sum_{\zeta} p^{\pi}(\zeta) \sum_{s,a,s'\sim \zeta}  \frac{\partial}{\partial \theta} g'_{\theta}(s,a,s')) \\
&= \sum_{s,a,s'} \left[ (\mu^E_{s,a,s'} - \mu^{\pi}_{s,a,s'}) \frac{\partial}{\partial \theta} g'_{\theta}(s,a,s') \right] 
\end{aligned}
\end{equation}

where we define
\begin{equation}
\begin{aligned}
g'_{\theta}(s,a,s') &= \log p(s'|s,a;\mu^{\pi}) + g_{\theta}(s,a), \\
\mu^E_{s,a,s'} &= \mu^E_{a} p(s'|s,a;\mu^E), \\ \mu^{\pi}_{s,a,s'} &= \mu^{\pi}_{a} p(s'|s,a;\mu^{\pi}),
\end{aligned}
\end{equation}
which represent visitation frequencies of state $s$, action $a$ and next state $s'$ in the expert trajectories $\mathcal{D}$ and a learned policy $\pi$.

Since we assumed that reward function influences dynamics only through pick-up probability which depends on visitation count $\mu^{\pi}$, we only consider $p(s'=o|s,a;\mu^{\pi})=\rho_s(\mu^{\pi})$ for transition probability:
\begin{equation}
\begin{aligned}
\frac{\partial L}{\partial \theta} &= \sum_{s,a} (\mu^E_a - \mu_a^{\pi}) \frac{\partial}{\partial \theta} g_{\theta}(s,a) \\
& + \sum_{s,a} (\mu^E_a \rho_a(\mu^E) - \mu_a^{\pi} \rho_a(\mu^{\pi})) \frac{\partial}{\partial \theta} \log \rho_a(\mu^{\pi}) \\
& + \sum_{s,a} (\mu^E_a (1 - \rho_a(\mu^E)) - \mu_a^{\pi} (1 - \rho_a(\mu^{\pi}))) \frac{\partial}{\partial \theta} \log (1 - \rho_a(\mu^{\pi})) \\
&= \sum_{s,a} \left[(\mu^E_a - \mu_a^{\pi})  \frac{\partial}{\partial \theta} g_{\theta}(s,a) 
+ \frac{\rho_a(\mu^E) - \rho_a(\mu^{\pi})}{1 - \rho_a(\mu^{\pi})} \mu^E_a \frac{\partial}{\partial \theta} \log \rho_a(\mu^{\pi}) \right]
\end{aligned}
\end{equation}
In Equation \ref{eq:pp}, suppose $\lambda << \mu + \sigma$, we can approximate pick-up probability by the Taylor expansion:

\begin{equation}
\begin{gathered}
\rho(\mu) = 1 - \exp \left( {\frac{\lambda}{\mu + \sigma}} \right) \approx \frac{\lambda}{\mu + \sigma} \\
\frac{\partial L}{\partial \theta} \approx \sum_{s,a} (\mu^E - \mu^{\pi}) \left( \frac{\partial g_{\theta}}{\partial \theta} + \rho_a(\mu^E) \left( \frac{\mu^E}{\mu^{\pi}+\sigma - \lambda} \right) \frac{\partial}{\partial \theta} \log \rho_a(\mu^{\pi}) \right)
\end{gathered}
\end{equation}

%% file: main.bbl
%%% -*-BibTeX-*-
%%% Do NOT edit. File created by BibTeX with style
%%% ACM-Reference-Format-Journals [18-Jan-2012].

\begin{thebibliography}{27}

%%% ====================================================================
%%% NOTE TO THE USER: you can override these defaults by providing
%%% customized versions of any of these macros before the \bibliography
%%% command.  Each of them MUST provide its own final punctuation,
%%% except for \shownote{}, \showDOI{}, and \showURL{}.  The latter two
%%% do not use final punctuation, in order to avoid confusing it with
%%% the Web address.
%%%
%%% To suppress output of a particular field, define its macro to expand
%%% to an empty string, or better, \unskip, like this:
%%%
%%% \newcommand{\showDOI}[1]{\unskip}   % LaTeX syntax
%%%
%%% \def \showDOI #1{\unskip}           % plain TeX syntax
%%%
%%% ====================================================================

\ifx \showCODEN    \undefined \def \showCODEN     #1{\unskip}     \fi
\ifx \showDOI      \undefined \def \showDOI       #1{#1}\fi
\ifx \showISBNx    \undefined \def \showISBNx     #1{\unskip}     \fi
\ifx \showISBNxiii \undefined \def \showISBNxiii  #1{\unskip}     \fi
\ifx \showISSN     \undefined \def \showISSN      #1{\unskip}     \fi
\ifx \showLCCN     \undefined \def \showLCCN      #1{\unskip}     \fi
\ifx \shownote     \undefined \def \shownote      #1{#1}          \fi
\ifx \showarticletitle \undefined \def \showarticletitle #1{#1}   \fi
\ifx \showURL      \undefined \def \showURL       {\relax}        \fi
% The following commands are used for tagged output and should be
% invisible to TeX
\providecommand\bibfield[2]{#2}
\providecommand\bibinfo[2]{#2}
\providecommand\natexlab[1]{#1}
\providecommand\showeprint[2][]{arXiv:#2}

\bibitem[\protect\citeauthoryear{Abbeel and Ng}{Abbeel and Ng}{2004}]%
        {abbeel2004apprenticeship}
\bibfield{author}{\bibinfo{person}{Pieter Abbeel} {and}
  \bibinfo{person}{Andrew~Y Ng}.} \bibinfo{year}{2004}\natexlab{}.
\newblock \showarticletitle{Apprenticeship learning via inverse reinforcement
  learning}. In \bibinfo{booktitle}{\emph{Proceedings of the twenty-first
  international conference on Machine learning}}. \bibinfo{pages}{1}.
\newblock


\bibitem[\protect\citeauthoryear{Altman, Boulogne, El-Azouzi, Jim{\'e}nez, and
  Wynter}{Altman et~al\mbox{.}}{2006}]%
        {altman2006survey}
\bibfield{author}{\bibinfo{person}{Eitan Altman}, \bibinfo{person}{Thomas
  Boulogne}, \bibinfo{person}{Rachid El-Azouzi}, \bibinfo{person}{Tania
  Jim{\'e}nez}, {and} \bibinfo{person}{Laura Wynter}.}
  \bibinfo{year}{2006}\natexlab{}.
\newblock \showarticletitle{A survey on networking games in
  telecommunications}.
\newblock \bibinfo{journal}{\emph{Computers \& Operations Research}}
  \bibinfo{volume}{33}, \bibinfo{number}{2} (\bibinfo{year}{2006}),
  \bibinfo{pages}{286--311}.
\newblock


\bibitem[\protect\citeauthoryear{Fu, Luo, and Levine}{Fu et~al\mbox{.}}{2017}]%
        {fu2017learning}
\bibfield{author}{\bibinfo{person}{Justin Fu}, \bibinfo{person}{Katie Luo},
  {and} \bibinfo{person}{Sergey Levine}.} \bibinfo{year}{2017}\natexlab{}.
\newblock \showarticletitle{Learning robust rewards with adversarial inverse
  reinforcement learning}.
\newblock \bibinfo{journal}{\emph{arXiv preprint arXiv:1710.11248}}
  (\bibinfo{year}{2017}).
\newblock


\bibitem[\protect\citeauthoryear{He and Shin}{He and Shin}{2019}]%
        {he2019spatio}
\bibfield{author}{\bibinfo{person}{Suining He} {and} \bibinfo{person}{Kang~G
  Shin}.} \bibinfo{year}{2019}\natexlab{}.
\newblock \showarticletitle{Spatio-temporal capsule-based reinforcement
  learning for mobility-on-demand network coordination}. In
  \bibinfo{booktitle}{\emph{The World Wide Web Conference}}.
  \bibinfo{pages}{2806--2813}.
\newblock


\bibitem[\protect\citeauthoryear{Ho and Ermon}{Ho and Ermon}{2016}]%
        {ho2016generative}
\bibfield{author}{\bibinfo{person}{Jonathan Ho} {and} \bibinfo{person}{Stefano
  Ermon}.} \bibinfo{year}{2016}\natexlab{}.
\newblock \showarticletitle{Generative adversarial imitation learning}. In
  \bibinfo{booktitle}{\emph{Advances in neural information processing
  systems}}. \bibinfo{pages}{4565--4573}.
\newblock


\bibitem[\protect\citeauthoryear{Hu, Wellman, et~al\mbox{.}}{Hu
  et~al\mbox{.}}{1998}]%
        {hu1998multiagent}
\bibfield{author}{\bibinfo{person}{Junling Hu}, \bibinfo{person}{Michael~P
  Wellman}, {et~al\mbox{.}}} \bibinfo{year}{1998}\natexlab{}.
\newblock \showarticletitle{Multiagent reinforcement learning: theoretical
  framework and an algorithm.}. In \bibinfo{booktitle}{\emph{ICML}},
  Vol.~\bibinfo{volume}{98}. Citeseer, \bibinfo{pages}{242--250}.
\newblock


\bibitem[\protect\citeauthoryear{Li, Qin, Jiao, Yang, Wang, Wang, Wu, and
  Ye}{Li et~al\mbox{.}}{2019}]%
        {li2019efficient}
\bibfield{author}{\bibinfo{person}{Minne Li}, \bibinfo{person}{Zhiwei Qin},
  \bibinfo{person}{Yan Jiao}, \bibinfo{person}{Yaodong Yang},
  \bibinfo{person}{Jun Wang}, \bibinfo{person}{Chenxi Wang},
  \bibinfo{person}{Guobin Wu}, {and} \bibinfo{person}{Jieping Ye}.}
  \bibinfo{year}{2019}\natexlab{}.
\newblock \showarticletitle{Efficient ridesharing order dispatching with mean
  field multi-agent reinforcement learning}. In \bibinfo{booktitle}{\emph{The
  World Wide Web Conference}}. \bibinfo{pages}{983--994}.
\newblock


\bibitem[\protect\citeauthoryear{Lin, Adams, and Beling}{Lin
  et~al\mbox{.}}{2019}]%
        {lin2019multi}
\bibfield{author}{\bibinfo{person}{Xiaomin Lin}, \bibinfo{person}{Stephen~C
  Adams}, {and} \bibinfo{person}{Peter~A Beling}.}
  \bibinfo{year}{2019}\natexlab{}.
\newblock \showarticletitle{Multi-agent inverse reinforcement learning for
  certain general-sum stochastic games}.
\newblock \bibinfo{journal}{\emph{Journal of Artificial Intelligence Research}}
   \bibinfo{volume}{66} (\bibinfo{year}{2019}), \bibinfo{pages}{473--502}.
\newblock


\bibitem[\protect\citeauthoryear{Lin, Beling, and Cogill}{Lin
  et~al\mbox{.}}{2017}]%
        {lin2017multiagent}
\bibfield{author}{\bibinfo{person}{Xiaomin Lin}, \bibinfo{person}{Peter~A
  Beling}, {and} \bibinfo{person}{Randy Cogill}.}
  \bibinfo{year}{2017}\natexlab{}.
\newblock \showarticletitle{Multiagent inverse reinforcement learning for
  two-person zero-sum games}.
\newblock \bibinfo{journal}{\emph{IEEE Transactions on Games}}
  \bibinfo{volume}{10}, \bibinfo{number}{1} (\bibinfo{year}{2017}),
  \bibinfo{pages}{56--68}.
\newblock


\bibitem[\protect\citeauthoryear{McKelvey and Palfrey}{McKelvey and
  Palfrey}{1995}]%
        {mckelvey1995quantal}
\bibfield{author}{\bibinfo{person}{Richard~D McKelvey} {and}
  \bibinfo{person}{Thomas~R Palfrey}.} \bibinfo{year}{1995}\natexlab{}.
\newblock \showarticletitle{Quantal response equilibria for normal form games}.
\newblock \bibinfo{journal}{\emph{Games and economic behavior}}
  \bibinfo{volume}{10}, \bibinfo{number}{1} (\bibinfo{year}{1995}),
  \bibinfo{pages}{6--38}.
\newblock


\bibitem[\protect\citeauthoryear{Mguni, Jennings, and de~Cote}{Mguni
  et~al\mbox{.}}{2018}]%
        {mguni2018decentralised}
\bibfield{author}{\bibinfo{person}{David Mguni}, \bibinfo{person}{Joel
  Jennings}, {and} \bibinfo{person}{Enrique~Munoz de Cote}.}
  \bibinfo{year}{2018}\natexlab{}.
\newblock \showarticletitle{Decentralised learning in systems with many, many
  strategic agents}.
\newblock \bibinfo{journal}{\emph{arXiv preprint arXiv:1803.05028}}
  (\bibinfo{year}{2018}).
\newblock


\bibitem[\protect\citeauthoryear{Mguni, Jennings, Macua, Sison, Ceppi, and
  de~Cote}{Mguni et~al\mbox{.}}{2019}]%
        {mguni2019coordinating}
\bibfield{author}{\bibinfo{person}{David Mguni}, \bibinfo{person}{Joel
  Jennings}, \bibinfo{person}{Sergio~Valcarcel Macua}, \bibinfo{person}{Emilio
  Sison}, \bibinfo{person}{Sofia Ceppi}, {and} \bibinfo{person}{Enrique~Munoz
  de Cote}.} \bibinfo{year}{2019}\natexlab{}.
\newblock \showarticletitle{Coordinating the crowd: Inducing desirable
  equilibria in non-cooperative systems}.
\newblock \bibinfo{journal}{\emph{arXiv preprint arXiv:1901.10923}}
  (\bibinfo{year}{2019}).
\newblock


\bibitem[\protect\citeauthoryear{Natarajan, Kunapuli, Judah, Tadepalli,
  Kersting, and Shavlik}{Natarajan et~al\mbox{.}}{2010}]%
        {natarajan2010multi}
\bibfield{author}{\bibinfo{person}{Sriraam Natarajan}, \bibinfo{person}{Gautam
  Kunapuli}, \bibinfo{person}{Kshitij Judah}, \bibinfo{person}{Prasad
  Tadepalli}, \bibinfo{person}{Kristian Kersting}, {and} \bibinfo{person}{Jude
  Shavlik}.} \bibinfo{year}{2010}\natexlab{}.
\newblock \showarticletitle{Multi-agent inverse reinforcement learning}. In
  \bibinfo{booktitle}{\emph{2010 Ninth International Conference on Machine
  Learning and Applications}}. IEEE, \bibinfo{pages}{395--400}.
\newblock


\bibitem[\protect\citeauthoryear{Oda and Joe-Wong}{Oda and Joe-Wong}{2018}]%
        {oda2018movi}
\bibfield{author}{\bibinfo{person}{Takuma Oda} {and} \bibinfo{person}{Carlee
  Joe-Wong}.} \bibinfo{year}{2018}\natexlab{}.
\newblock \showarticletitle{MOVI: A model-free approach to dynamic fleet
  management}. In \bibinfo{booktitle}{\emph{IEEE INFOCOM 2018-IEEE Conference
  on Computer Communications}}. IEEE, \bibinfo{pages}{2708--2716}.
\newblock


\bibitem[\protect\citeauthoryear{Pomerleau}{Pomerleau}{1991}]%
        {pomerleau1991efficient}
\bibfield{author}{\bibinfo{person}{Dean~A Pomerleau}.}
  \bibinfo{year}{1991}\natexlab{}.
\newblock \showarticletitle{Efficient training of artificial neural networks
  for autonomous navigation}.
\newblock \bibinfo{journal}{\emph{Neural computation}} \bibinfo{volume}{3},
  \bibinfo{number}{1} (\bibinfo{year}{1991}), \bibinfo{pages}{88--97}.
\newblock


\bibitem[\protect\citeauthoryear{Qu, Zhu, Liu, Liu, and Xiong}{Qu
  et~al\mbox{.}}{2014}]%
        {qu2014cost}
\bibfield{author}{\bibinfo{person}{Meng Qu}, \bibinfo{person}{Hengshu Zhu},
  \bibinfo{person}{Junming Liu}, \bibinfo{person}{Guannan Liu}, {and}
  \bibinfo{person}{Hui Xiong}.} \bibinfo{year}{2014}\natexlab{}.
\newblock \showarticletitle{A cost-effective recommender system for taxi
  drivers}. In \bibinfo{booktitle}{\emph{Proceedings of the 20th ACM SIGKDD
  international conference on Knowledge discovery and data mining}}.
  \bibinfo{pages}{45--54}.
\newblock


\bibitem[\protect\citeauthoryear{Sutton, Precup, and Singh}{Sutton
  et~al\mbox{.}}{1999}]%
        {sutton1999between}
\bibfield{author}{\bibinfo{person}{Richard~S Sutton}, \bibinfo{person}{Doina
  Precup}, {and} \bibinfo{person}{Satinder Singh}.}
  \bibinfo{year}{1999}\natexlab{}.
\newblock \showarticletitle{Between MDPs and semi-MDPs: A framework for
  temporal abstraction in reinforcement learning}.
\newblock \bibinfo{journal}{\emph{Artificial intelligence}}
  \bibinfo{volume}{112}, \bibinfo{number}{1-2} (\bibinfo{year}{1999}),
  \bibinfo{pages}{181--211}.
\newblock


\bibitem[\protect\citeauthoryear{Sykora, Ren, and Urtasun}{Sykora
  et~al\mbox{.}}{2020}]%
        {sykora2020multi}
\bibfield{author}{\bibinfo{person}{Quinlan Sykora}, \bibinfo{person}{Mengye
  Ren}, {and} \bibinfo{person}{Raquel Urtasun}.}
  \bibinfo{year}{2020}\natexlab{}.
\newblock \showarticletitle{Multi-agent routing value iteration network}.
\newblock \bibinfo{journal}{\emph{arXiv preprint arXiv:2007.05096}}
  (\bibinfo{year}{2020}).
\newblock


\bibitem[\protect\citeauthoryear{Tang, Qin, Zhang, Wang, Xu, Ma, Zhu, and
  Ye}{Tang et~al\mbox{.}}{2019}]%
        {tang2019deep}
\bibfield{author}{\bibinfo{person}{Xiaocheng Tang}, \bibinfo{person}{Zhiwei
  Qin}, \bibinfo{person}{Fan Zhang}, \bibinfo{person}{Zhaodong Wang},
  \bibinfo{person}{Zhe Xu}, \bibinfo{person}{Yintai Ma},
  \bibinfo{person}{Hongtu Zhu}, {and} \bibinfo{person}{Jieping Ye}.}
  \bibinfo{year}{2019}\natexlab{}.
\newblock \showarticletitle{A deep value-network based approach for
  multi-driver order dispatching}. In \bibinfo{booktitle}{\emph{Proceedings of
  the 25th ACM SIGKDD international conference on knowledge discovery \& data
  mining}}. \bibinfo{pages}{1780--1790}.
\newblock


\bibitem[\protect\citeauthoryear{Waugh, Ziebart, and Bagnell}{Waugh
  et~al\mbox{.}}{2013}]%
        {waugh2013computational}
\bibfield{author}{\bibinfo{person}{Kevin Waugh}, \bibinfo{person}{Brian~D
  Ziebart}, {and} \bibinfo{person}{J~Andrew Bagnell}.}
  \bibinfo{year}{2013}\natexlab{}.
\newblock \showarticletitle{Computational rationalization: The inverse
  equilibrium problem}.
\newblock \bibinfo{journal}{\emph{arXiv preprint arXiv:1308.3506}}
  (\bibinfo{year}{2013}).
\newblock


\bibitem[\protect\citeauthoryear{Xu, Li, Guan, Zhang, Li, Nan, Liu, Bian, and
  Ye}{Xu et~al\mbox{.}}{2018}]%
        {xu2018large}
\bibfield{author}{\bibinfo{person}{Zhe Xu}, \bibinfo{person}{Zhixin Li},
  \bibinfo{person}{Qingwen Guan}, \bibinfo{person}{Dingshui Zhang},
  \bibinfo{person}{Qiang Li}, \bibinfo{person}{Junxiao Nan},
  \bibinfo{person}{Chunyang Liu}, \bibinfo{person}{Wei Bian}, {and}
  \bibinfo{person}{Jieping Ye}.} \bibinfo{year}{2018}\natexlab{}.
\newblock \showarticletitle{Large-scale order dispatch in on-demand
  ride-hailing platforms: A learning and planning approach}. In
  \bibinfo{booktitle}{\emph{Proceedings of the 24th ACM SIGKDD International
  Conference on Knowledge Discovery \& Data Mining}}.
  \bibinfo{pages}{905--913}.
\newblock


\bibitem[\protect\citeauthoryear{Youn, Gastner, and Jeong}{Youn
  et~al\mbox{.}}{2008}]%
        {youn2008price}
\bibfield{author}{\bibinfo{person}{Hyejin Youn}, \bibinfo{person}{Michael~T
  Gastner}, {and} \bibinfo{person}{Hawoong Jeong}.}
  \bibinfo{year}{2008}\natexlab{}.
\newblock \showarticletitle{Price of anarchy in transportation networks:
  efficiency and optimality control}.
\newblock \bibinfo{journal}{\emph{Physical review letters}}
  \bibinfo{volume}{101}, \bibinfo{number}{12} (\bibinfo{year}{2008}),
  \bibinfo{pages}{128701}.
\newblock


\bibitem[\protect\citeauthoryear{Yu, Song, and Ermon}{Yu et~al\mbox{.}}{2019}]%
        {yu2019multi}
\bibfield{author}{\bibinfo{person}{Lantao Yu}, \bibinfo{person}{Jiaming Song},
  {and} \bibinfo{person}{Stefano Ermon}.} \bibinfo{year}{2019}\natexlab{}.
\newblock \showarticletitle{Multi-agent adversarial inverse reinforcement
  learning}.
\newblock \bibinfo{journal}{\emph{arXiv preprint arXiv:1907.13220}}
  (\bibinfo{year}{2019}).
\newblock


\bibitem[\protect\citeauthoryear{Yuan, Zheng, Zhang, and Xie}{Yuan
  et~al\mbox{.}}{2012}]%
        {yuan2012t}
\bibfield{author}{\bibinfo{person}{Nicholas~Jing Yuan}, \bibinfo{person}{Yu
  Zheng}, \bibinfo{person}{Liuhang Zhang}, {and} \bibinfo{person}{Xing Xie}.}
  \bibinfo{year}{2012}\natexlab{}.
\newblock \showarticletitle{T-finder: A recommender system for finding
  passengers and vacant taxis}.
\newblock \bibinfo{journal}{\emph{IEEE Transactions on knowledge and data
  engineering}} \bibinfo{volume}{25}, \bibinfo{number}{10}
  (\bibinfo{year}{2012}), \bibinfo{pages}{2390--2403}.
\newblock


\bibitem[\protect\citeauthoryear{Zhang and Pavone}{Zhang and Pavone}{2016}]%
        {zhang2016control}
\bibfield{author}{\bibinfo{person}{Rick Zhang} {and} \bibinfo{person}{Marco
  Pavone}.} \bibinfo{year}{2016}\natexlab{}.
\newblock \showarticletitle{Control of robotic mobility-on-demand systems: a
  queueing-theoretical perspective}.
\newblock \bibinfo{journal}{\emph{The International Journal of Robotics
  Research}} \bibinfo{volume}{35}, \bibinfo{number}{1-3}
  (\bibinfo{year}{2016}), \bibinfo{pages}{186--203}.
\newblock


\bibitem[\protect\citeauthoryear{Ziebart, Maas, Bagnell, and Dey}{Ziebart
  et~al\mbox{.}}{2008a}]%
        {ziebart2008maximum}
\bibfield{author}{\bibinfo{person}{Brian~D Ziebart}, \bibinfo{person}{Andrew~L
  Maas}, \bibinfo{person}{J~Andrew Bagnell}, {and} \bibinfo{person}{Anind~K
  Dey}.} \bibinfo{year}{2008}\natexlab{a}.
\newblock \showarticletitle{Maximum entropy inverse reinforcement learning.}.
  In \bibinfo{booktitle}{\emph{Aaai}}, Vol.~\bibinfo{volume}{8}. Chicago, IL,
  USA, \bibinfo{pages}{1433--1438}.
\newblock


\bibitem[\protect\citeauthoryear{Ziebart, Maas, Dey, and Bagnell}{Ziebart
  et~al\mbox{.}}{2008b}]%
        {ziebart2008navigate}
\bibfield{author}{\bibinfo{person}{Brian~D Ziebart}, \bibinfo{person}{Andrew~L
  Maas}, \bibinfo{person}{Anind~K Dey}, {and} \bibinfo{person}{J~Andrew
  Bagnell}.} \bibinfo{year}{2008}\natexlab{b}.
\newblock \showarticletitle{Navigate like a cabbie: Probabilistic reasoning
  from observed context-aware behavior}. In
  \bibinfo{booktitle}{\emph{Proceedings of the 10th international conference on
  Ubiquitous computing}}. \bibinfo{pages}{322--331}.
\newblock


\end{thebibliography}
